\documentclass[conference,10pt]{IEEEtran}
\IEEEoverridecommandlockouts                              

\usepackage{times}

\usepackage[bookmarks=true,colorlinks=true,urlcolor=blue,citecolor=blue]{hyperref}

\usepackage{caption}
\usepackage{comment}
\usepackage{siunitx}
\usepackage{relsize}
\usepackage{ifthen}
\usepackage[colorinlistoftodos]{todonotes}
\usepackage{wrapfig}
\usepackage[caption=false]{subfig}

\usepackage[vlined,ruled,linesnumbered]{algorithm2e}
\usepackage{graphics} 
\usepackage{rotating}
\usepackage{color}
\usepackage{enumerate}
\usepackage[T1]{fontenc}
\usepackage{psfrag}
\usepackage{epsfig} 
\usepackage{booktabs}
\usepackage{graphicx,url}
\usepackage{multirow}
\usepackage{array}
\usepackage{latexsym}
\usepackage{amsfonts}
\usepackage{amsmath}
\usepackage{amssymb}
\usepackage{xstring}
\usepackage[noend]{algorithmic}
\usepackage{multirow}
\usepackage{xcolor}
\usepackage{prettyref}
\usepackage{flexisym}
\usepackage{bigdelim}
\usepackage{breqn} 
\usepackage{listings}
\let\labelindent\relax
\usepackage{enumitem}
\usepackage{xspace}
\usepackage{bm}
\graphicspath{{./figures/}}
\usepackage{tikz}
\usetikzlibrary{matrix,calc}

\usepackage{cleveref}

\usepackage{mdwlist}

\let\itemize\undefined
\makecompactlist{itemize}{stditemize}

\newrefformat{prob}{Problem\,\ref{#1}}
\newrefformat{def}{Definition\,\ref{#1}}
\newrefformat{sec}{Section\,\ref{#1}}
\newrefformat{sub}{Section\,\ref{#1}}
\newrefformat{prop}{Proposition\,\ref{#1}}
\newrefformat{app}{Appendix\,\ref{#1}}
\newrefformat{alg}{Algorithm\,\ref{#1}}
\newrefformat{cor}{Corollary\,\ref{#1}}
\newrefformat{thm}{Theorem\,\ref{#1}}
\newrefformat{lem}{Lemma\,\ref{#1}}
\newrefformat{fig}{Fig.\,\ref{#1}}
\newrefformat{tab}{Table\,\ref{#1}}

\newcommand{\bdmath}{\begin{dmath}}
\newcommand{\edmath}{\end{dmath}}
\newcommand{\beq}{\begin{equation}}
\newcommand{\eeq}{\end{equation}}
\newcommand{\bdm}{\begin{displaymath}}
\newcommand{\edm}{\end{displaymath}}
\newcommand{\bea}{\begin{eqnarray}}
\newcommand{\eea}{\end{eqnarray}}
\newcommand{\beal}{\beq \begin{array}{ll}}
\newcommand{\eeal}{\end{array} \eeq}
\newcommand{\beas}{\begin{eqnarray*}}
\newcommand{\eeas}{\end{eqnarray*}}
\newcommand{\ba}{\begin{array}}
\newcommand{\ea}{\end{array}}
\newcommand{\bit}{\begin{itemize}}
\newcommand{\eit}{\end{itemize}}
\newcommand{\ben}{\begin{enumerate}}
\newcommand{\een}{\end{enumerate}}

\newcommand{\etal}{\emph{et~al.}\xspace}

\newcommand{\eg}{\emph{e.g.,}\xspace}
\newcommand{\ie}{\emph{i.e.,}\xspace}

\newcommand{\hide}[1]{}

\newcommand{\hiddenText}{{\color{gray} hidden text.}}
\newcommand{\hideWithText}[1]{\hiddenText}

\newcommand{\pinv}{^\dag}

\newcommand{\scenario}[1]{{\smaller \sf#1}\xspace}

\newcommand{\Outdoor}{\scenario{Campus-Outdoor}}
\newcommand{\Tunnels}{\scenario{Campus-Tunnels}}
\newcommand{\Hybrid}{\scenario{Campus-Hybrid}}

\newcommand{\blue}[1]{{\color{blue}#1}}

\newcommand{\linkToPdf}[1]{\href{#1}{\blue{(pdf)}}}
\newcommand{\linkToPpt}[1]{\href{#1}{\blue{(ppt)}}}
\newcommand{\linkToCode}[1]{\href{#1}{\blue{(code)}}}
\newcommand{\linkToWeb}[1]{\href{#1}{\blue{(web)}}}
\newcommand{\linkToVideo}[1]{\href{#1}{\blue{(video)}}}
\newcommand{\linkToMedia}[1]{\href{#1}{\blue{(media)}}}
\newcommand{\award}[1]{\xspace} 

\newcommand{\kimeraMulti}{\scenario{Kimera-Multi}} 

\usepackage[numbers,sort&compress]{natbib}
\begin{document}

\title{\LARGE \bf
	Resilient and Distributed Multi-Robot Visual SLAM:
	\\Datasets, Experiments, and Lessons Learned
}

\author{
	Yulun Tian$^{1\pinv}$, 
	Yun Chang$^{1\pinv}$, 
	Long Quang$^{2}$, 
	Arthur Schang$^{3}$,
	Carlos Nieto-Granda$^{2}$,
	Jonathan P. How$^{1}$, 
	Luca Carlone$^{1}$
	\thanks{{*This work was supported in part by ARL DCIST
			under Cooperative Agreement Number W911NF-17-2-0181, 
			in part by ONR under BRC Award N000141712072, 
			and in part by MathWorks.
		}}
	\thanks{$\pinv$ Equal contribution.}
	\thanks{$^{1}$Y. Tian, Y. Chang, J.~P.~How, L.~Carlone are with the Department of Aeronautics and Astronautics, Massachusetts Institute of Technology, 77 Massachusetts Ave, Cambridge, MA 02139, USA.
		{\tt\small \{yulun,yunchang,jhow,lcarlone\}@mit.edu}}%
	\thanks{$^{2}$L.~Quang and C.~Nieto-Granda are with U.S. Army Combat Capabilities Development
		Command, Army Research Laboratory, Adelphi, MD 20783, USA.
		{\tt\small \{long.p.quang.civ, carlos.p.nieto2.civ\}@army.mil}}%
	\thanks{$^{3}$A.~Schang is with Parsons Corporation, Centreville, VA 20120, USA.
		{\tt\small arthur.schang@parsons.com}}%
	\thanks{
		The authors would like to thank 
		Jason Hughes, 
		Pratheek Manjunath,
		Varun Murali,
		Mason Peterson,
		Aaron Ray,
		and Nathan Hughes
		for assistance with outdoor experiments and data collection,
		Kaveh Fathian,
		Jared Strader,
		Lakshay Sharma for hardware supports,
		Kevin Garcia for assistance with the dataset release,
		Jonathan Fink for an early implementation of remote topic manager,
		Prof.~Thomas Herring and his group for discussion about RTK GPS,
		and 
		Anthony Zolnik for logistics support.
		}
}

\maketitle



\begin{abstract}

This paper revisits \kimeraMulti, a distributed multi-robot Simultaneous Localization and Mapping (SLAM) system,
towards the goal of deployment in the real world.
In particular, this paper has three main contributions.
First, we describe improvements to \kimeraMulti to make it resilient to large-scale real-world deployments,
with particular emphasis on handling intermittent and unreliable communication.
Second, we collect and release challenging multi-robot benchmarking datasets
obtained during live experiments conducted on the MIT campus,
with accurate reference trajectories and maps
for evaluation. 
The datasets include up to 8 robots traversing long distances (up to 8 km)
and feature many challenging elements such as severe visual ambiguities (\eg in underground tunnels and hallways), 
mixed indoor and outdoor trajectories with different lighting conditions, and dynamic entities (\eg pedestrians and cars).
Lastly, we evaluate the resilience of \kimeraMulti under different communication scenarios,
and provide a quantitative comparison with a centralized baseline system.
Based on the results from both live experiments and subsequent analysis, we discuss the strengths and weaknesses of \kimeraMulti,
and suggest future directions for both algorithm and system design. 
We release the source code of \kimeraMulti and all datasets
to facilitate further research towards the reliable real-world deployment of multi-robot SLAM systems.

\end{abstract}

\bstctlcite{bst_control}

\IEEEpeerreviewmaketitle


\section{Introduction}
\label{sec:intro}

Collaborative Simultaneous Localization and Mapping (CSLAM) is a fundamental capability that enables multiple robots to operate in GPS-denied environments.
Recently, significant progress has been made towards designing more efficient and robust CSLAM algorithms and systems
supporting different sensor modalities and communication architectures~\cite{Lajoie21Survey}.
In particular, the DARPA  Subterranean Challenge~\cite{SubT}
has seen successful deployments of lidar-centric CSLAM systems (\eg~\cite{Chang22ral-LAMP2,Tranzatto22Cerberus,Ebadi22arxiv-surveySLAMSubt}) 
that enable teams of 
robots to explore extreme underground environments. 

\begin{figure}[!t]
	\centering
	\includegraphics[trim={0 0 0 0},clip, width=1.0\columnwidth]{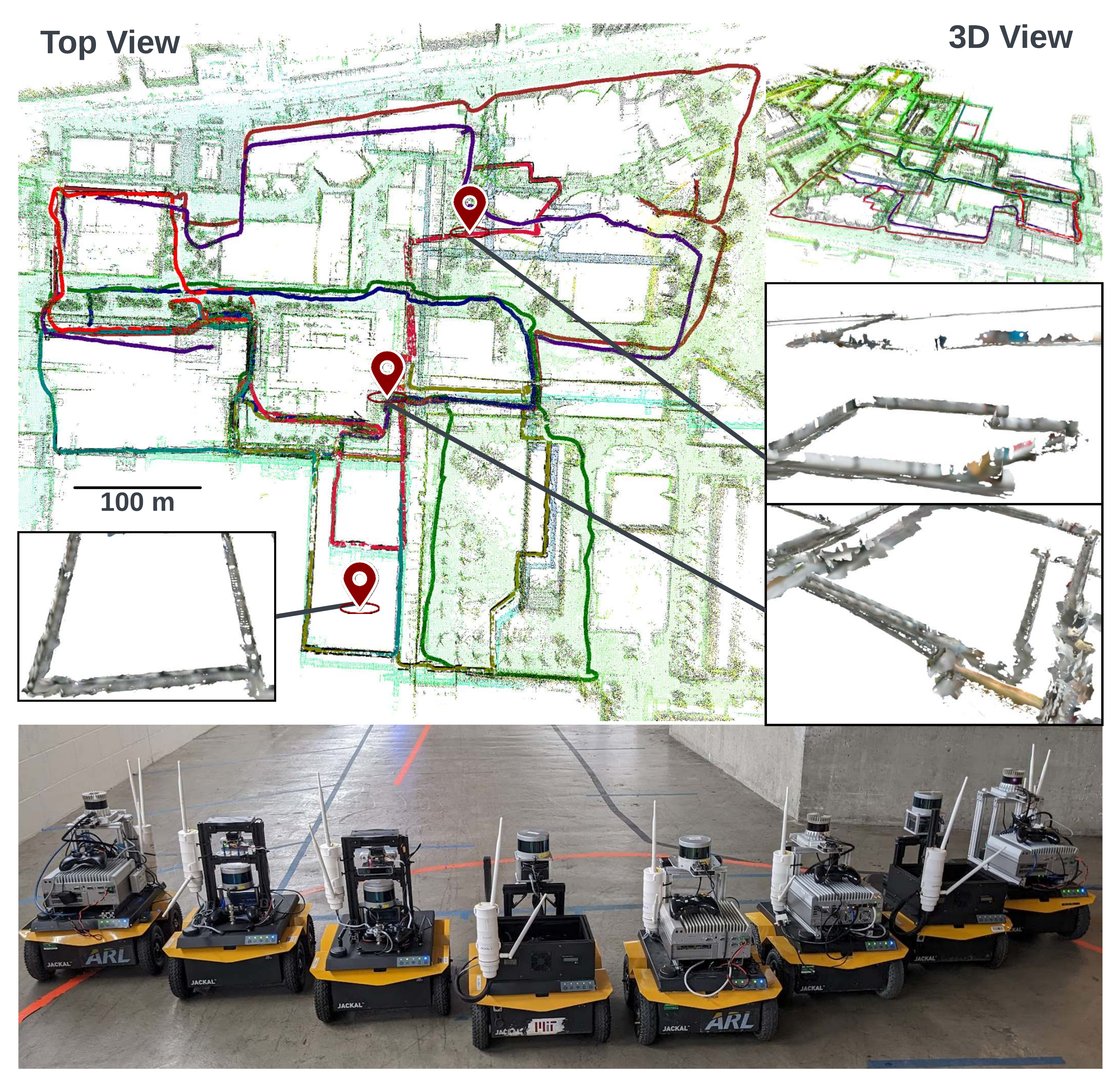}
	\caption{\kimeraMulti trajectories and meshes overlaid on top of the reference point-cloud map from 
		the \Hybrid experiment where 8 robots traverse indoor-outdoor scenes covering a total of $7785$m.\label{fig:cover}}
	\vspace{-8mm}
\end{figure}

Compared to lidar-centric CSLAM,
evaluations of \emph{vision-based} CSLAM systems
in large-scale and complex environments have been relatively lacking.
In an effort to bridge this gap, 
we conduct live experiments and present an in-depth analysis of a state-of-the-art visual-inertial CSLAM system, 
\kimeraMulti~\cite{Chang21icra-KimeraMulti,Tian21tro-KimeraMulti},
on challenging urban scenarios involving 6-8 robots traversing up to nearly 8 kilometers combined; see Fig.~\ref{fig:cover}.
Different from prior works on vision-based CSLAM systems with large-scale experiments (\eg \cite{Schmuck2021Covins}),
we focus on a \emph{fully distributed} CSLAM framework,
and our test scenarios are more diverse (\eg including both indoor and outdoor portions in a single trajectory)
and involve more dynamic entities (\eg moving cars and pedestrians captured by front-facing cameras).
These challenging experiments allow us to assess
(i)  the \emph{scalability} of our system to support many robots operating in large environments for extended periods of time,
and 
(ii) \emph{resilience} against real-world operational challenges such as intermittent communication, dynamic environments, and hardware failures.
We show that \kimeraMulti is able to achieve accurate results and offers additional flexibility compared to a centralized system,
and discuss new insights and lessons learned from our field tests and subsequent analysis.

{\bf Contributions.}
This paper describes our experimental efforts to evaluate the resiliency of \kimeraMulti in the real world.
In particular, we have three goals.
First, we describe the improvements we made to enable real-world deployment of \kimeraMulti.
Then, we describe the live experiments conducted on the MIT campus
along with the challenging large-scale benchmarking datasets compiled from data recorded during the live experiments.
Finally, we provide quantitative results from controlled experiments and discuss lessons learned from live field tests.
We release the source code of \kimeraMulti\footnote{\url{https://github.com/MIT-SPARK/Kimera-Multi}} 
and all datasets\footnote{\url{https://github.com/MIT-SPARK/Kimera-Multi-Data}} together with accurate reference trajectories and point-cloud maps
to facilitate further research in this area.


\section{Related Work}
\label{sec:relatedWork}

Collaborative SLAM (CSLAM) has recently received significant attention.
Recent works have 
investigated the CSLAM \emph{front-end} for efficiently establishing inter-robot loop closures~\cite{Cieslewski18icra,Giamou18icra,tian19resource},
and the \emph{back-end} for solving the underlying geometric estimation problem in a distributed fashion~\cite{Cunningham10iros,Choudhary17ijrr-distributedPGO3D,tian2019distributed,tian2020asynchronous,Fan21Majorization,Murai22RobotWeb}. 
The reader is referred to the recent survey~\cite{Lajoie21Survey} for detailed descriptions of these algorithmic advances.

{\bf CSLAM Systems.}
Existing CSLAM systems can be categorized based on whether they implement a \emph{centralized} or \emph{distributed} architecture.
CCM-SLAM~\cite{Schmuck18CCM} is a well-established centralized system for visual-inertial CSLAM, in which a central server is responsible for multi-robot map management, fusion, and optimization.
COVINS~\cite{Schmuck2021Covins,Patel23Covinsg} further extends~\cite{Schmuck18CCM} and is demonstrated to scale to 
12 robots.
CVIDS~\cite{Zhang22cvids} is another recent centralized CSLAM system that produces a dense global TSDF map.
LAMP~\cite{Ebadi20icra-LAMP,Chang22ral-LAMP2} is a state-of-the-art centralized system for lidar-centric CSLAM
and includes a loop closure prioritization module and an outlier-robust pose graph optimization (PGO) module based on graduated non-convexity (GNC)~\cite{Yang20ral-GNC}.
While centralized systems offer great accuracy and ease of data management, 
they often require a stable connection with the server and are susceptible to a single point of failure.

Distributed systems seek to alleviate the aforementioned limitations by removing the dependence on the central server.
Zhang \etal~\cite{Zhang18Distributed} develop a distributed system for monocular-only CSLAM, where each robot performs global map merging onboard.
Cieslewski~\etal~\cite{Cieslewski18icra} and DOOR-SLAM~\cite{Lajoie20ral-doorSLAM} apply \emph{distributed} PGO using the distributed Gauss-Seidel (DGS) method~\cite{Choudhary17ijrr-distributedPGO3D}.
DOOR-SLAM~\cite{Lajoie20ral-doorSLAM} further employs Pairwise Consistency Maximization (PCM)~\cite{Mangelson18icra} to reject outlier inter-robot loop closures.
In our previous works~\cite{Chang21icra-KimeraMulti,Tian21tro-KimeraMulti},
we developed \kimeraMulti, which includes an outlier-robust back-end based on distributed GNC that outperforms PCM in terms of accuracy.
$D^2$SLAM~\cite{Xu22D2slam} is a recent system that applies distributed and asynchronous optimization on multi-robot VIO and PGO.
In parallel, Huang~\etal~\cite{Huang21Disco} and Zhong~\etal~\cite{Zhong22DCL} also develop distributed systems for lidar-based CSLAM.
Swarm-SLAM~\cite{Lajoie23Swarm} is a very recent open-source system that supports both visual and lidar sensors.
Building on a spectral sparsification method for single-robot SLAM~\cite{Doherty22MAC},
Swarm-SLAM prioritizes inter-robot loop closures by
selecting candidates that maximize the algebraic connectivity of the multi-robot measurement graph.
In contrast to the distributed back-ends used by previous works~\cite{Cieslewski18icra,Lajoie20ral-doorSLAM,Tian21tro-KimeraMulti},
Swarm-SLAM implements a protocol that 
dynamically elects a leader among the connected robots to solve the full multi-robot PGO problem.

{\bf CSLAM Datasets.}
Apart from recent system works, several research groups have also contributed new large-scale CSLAM datasets.
The NeBula~\cite{Chang22ral-LAMP2} and CERBERUS~\cite{Tranzatto22Cerberus} datasets are collected during the 
recent DARPA Subterranean Challenge.
GRACO~\cite{Zhu23Graco} includes multiple ground and aerial sequences for evaluating CSLAM using heterogeneous platforms.
S3E~\cite{Feng22S3E} is a collection of CSLAM datasets that includes multiple indoor and outdoor trajectory designs with varying difficulties for 3 robots.
M2DGR~\cite{Yin21M2dgr} are datasets collected by a single robot that traverses diverse scenarios,
and contains multiple sequences in the same environments.


\section{System Description}
\label{sec:system}
\begin{figure}[!t]
    \centering
    \includegraphics[trim={0 0 0 0},clip, width=1.0\columnwidth]{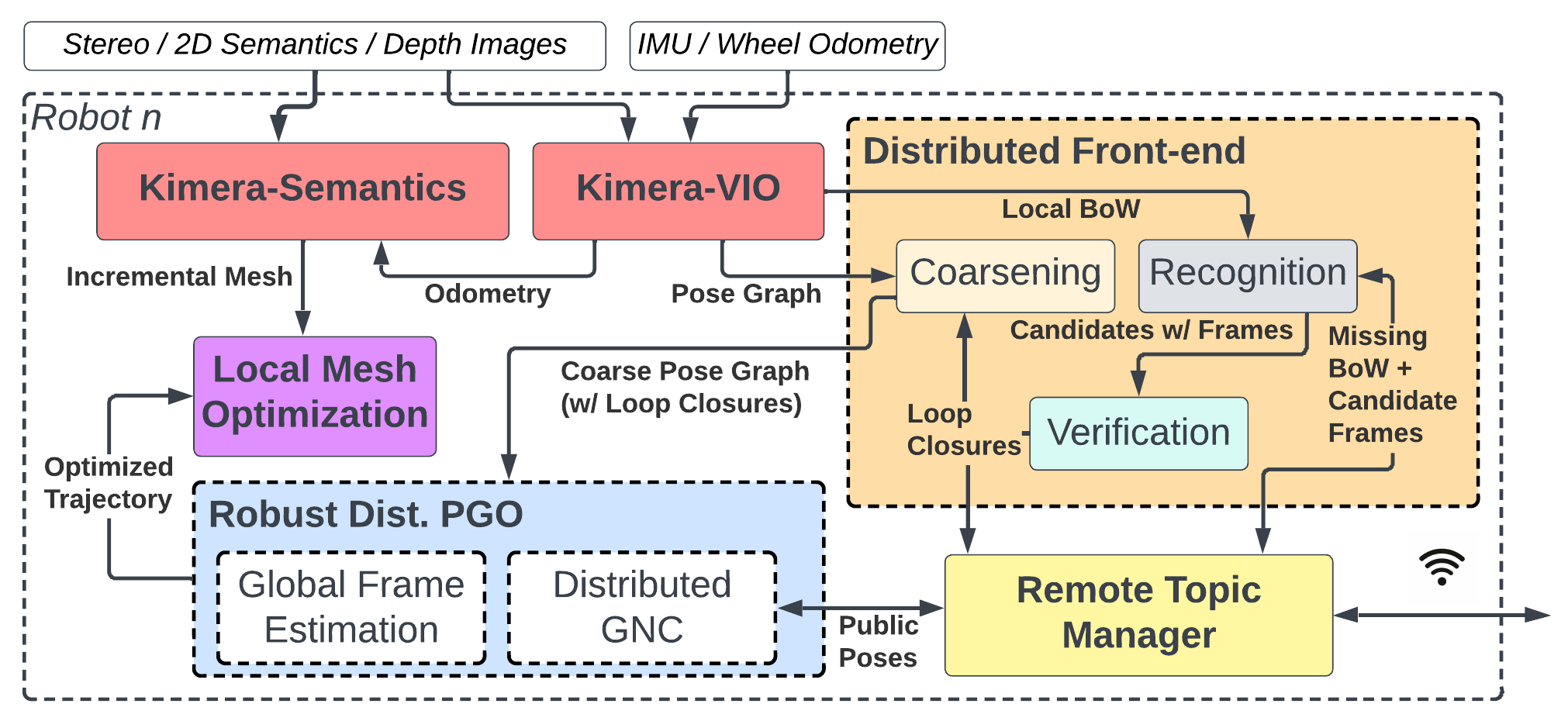}
    \caption{Modules on a robot in the distributed multi-robot system and the message flow from sensor data to trajectory and reconstruction.\label{fig:system}}
    \vspace{-7mm}
\end{figure}

The \kimeraMulti system~\cite{Chang21icra-KimeraMulti,Tian21tro-KimeraMulti} is summarized in Fig.~\ref{fig:system}.
Each robot runs an onboard system using the Robot Operating System (ROS)~\cite{Quigley09icra-ros}.
Inter-robot communication is {performed in a \emph{peer-to-peer} manner} using a lightweight 
communication layer on top of the UDP protocol.
Kimera-VIO and Kimera-Semantics~\cite{Rosinol21ijrr-Kimera} provide the
odometric pose estimates and a reconstructed 3D mesh.
The distributed \emph{front-end} detects inter-robot loop closures by 
communicating visual Bag-of-Words (BoW) vectors and selected keyframes that contain keypoints and descriptors for geometric verification.
The front-end is also responsible for incorporating the odometry and loop closures into a \emph{coarsened} pose graph.
The distributed \emph{back-end} periodically optimizes the coarse pose graph using robust distributed optimization.
Lastly, the optimized trajectory is used 
by each robot to correct its local 3D mesh.

{\bf Distributed Front-end.}
{Whenever two robots can communicate, one of the robots executes the \kimeraMulti front-end to detect inter-robot loop closures.\footnote{In our implementation, between each pair of robots, the robot with a smaller ID is designated to run the front-end.}}
The front-end consists of three main components: 
\emph{place recognition}, \emph{geometric verification}, and \emph{pose graph coarsening}.

{The \emph{place recognition} component subscribes to BoW vectors from the other robot, and finds matches by searching the local database of past BoW vectors.}
To avoid a high bandwidth usage, we perform an optional downsampling and only transmit every $n_b$th BoW vector (default $n_b=3$).
A candidate loop closure is identified if the normalized similarity score is higher than a threshold $\alpha$ (default 0.5).
In order to operate robustly under 
sporadic communication,
compared to our previous system~\cite{Tian21tro-KimeraMulti}, 
our latest implementation tolerates \emph{out-of-order} arrival of BoW vectors.
Each robot also periodically examines its local database 
for missing BoW vectors, 
and publishes a request to the connected robot that has the most missing BoW vectors.

The \emph{geometric verification} component processes the queue of candidate loop closures as in \cite{Tian21tro-KimeraMulti}. 
For each candidate inter-robot loop closure, a request is transmitted to the other robot to send its corresponding visual keyframe, which contains 3D keypoints and ORB descriptors. 
This robot then performs standard descriptor matching and computes the underlying relative transformation using monocular and stereo RANSAC.
The estimated transformation is sent back to the other robot to be included in its local pose graph.

To prevent the rapid growth of the multi-robot pose graph, 
we add a \emph{pose graph coarsening} component that subscribes to the local pose graph and loop closures,
and then reduces the graph by aggregating pose variables within a specified distance $d$ (default $d=2$m).
{In our experiments, the coarsened pose graph is on average 90\% smaller than the original pose graph,
which enables more efficient optimization in our back-end.}

{\bf Distributed Robust Pose Graph Optimization.}
At the start of each mission, robots initialize their trajectory and map estimates in their respective local frames.
Once the front-end detects inter-robot loop closures, robots solve distributed PGO with the two-stage distributed back-end developed in our previous work~\cite{Tian21tro-KimeraMulti}.
In the first stage, each robot estimates its relative transformation to the global frame\footnote{Without loss of generality, we assign the global reference frame by anchoring the initial pose of the first robot at identity.} 
by locally solving a robust single pose averaging problem using GNC.
The estimated transformation is accepted if the solution is supported by at least 3 inlier loop closures.

Once robots are initialized in the global frame, 
the second stage solves robust PGO via distributed GNC~\cite[Sec.~V]{Tian21tro-KimeraMulti}.
Instead of waiting for all robots to be connected, our latest system enables concurrent PGO within multiple clusters of connected robots.
To maintain consistency across clusters, a prior over the global reference frame is added to each cluster.\footnote{The prior is obtained from the last round of distributed optimization involving the first robot (who we assign as the global reference frame).
	Within each cluster, the prior is added to the robot with the smallest ID.}
Within each cluster, robots iteratively refine their trajectory estimates by performing local optimization steps and communicating public poses (\ie poses associated with inter-robot loop closures) with others \cite{tian2019distributed}.
For this purpose, we assume that robots within the same cluster can reach each other (either via direct links or additional routing), so that they can transmit their public poses to the intended destination and maintain synchronization during distributed optimization.
The current round of PGO is terminated once the relative changes of all robots' translation estimates are less than a threshold $\epsilon_{\text{rel}}$ (default 0.2m).
Compared to our previous work~\cite{Tian21tro-KimeraMulti}, our latest system better handles sporadic communication and situations when only a subset of the robots are connected.

{\bf Inter-Robot Communication.}
Our latest system implements a \emph{remote topic manager} module to handle communication between robots.
This module closely integrates with the ROS publish-subscribe paradigm 
and manages incoming and outgoing ROS messages with other robots.
The remote topic manager also keeps track of currently connected robots,
and initiates new connections when others are within communication range.
The remote topic manager is implemented using the open-source ENet library,\footnote{{\url{http://enet.bespin.org/}}}
which provides lightweight communication using UDP and supports reliable, in-order transmission of selected data streams.
Lastly, ENet also provides diagnostic statistics such as delay and packet loss that are 
helpful for evaluating communication performance.


\section{Datasets}
\label{sec:dataset}
\begin{figure}[!t]
    \centering
    \includegraphics[trim={20 20 10 20},clip, width=1.0\columnwidth]{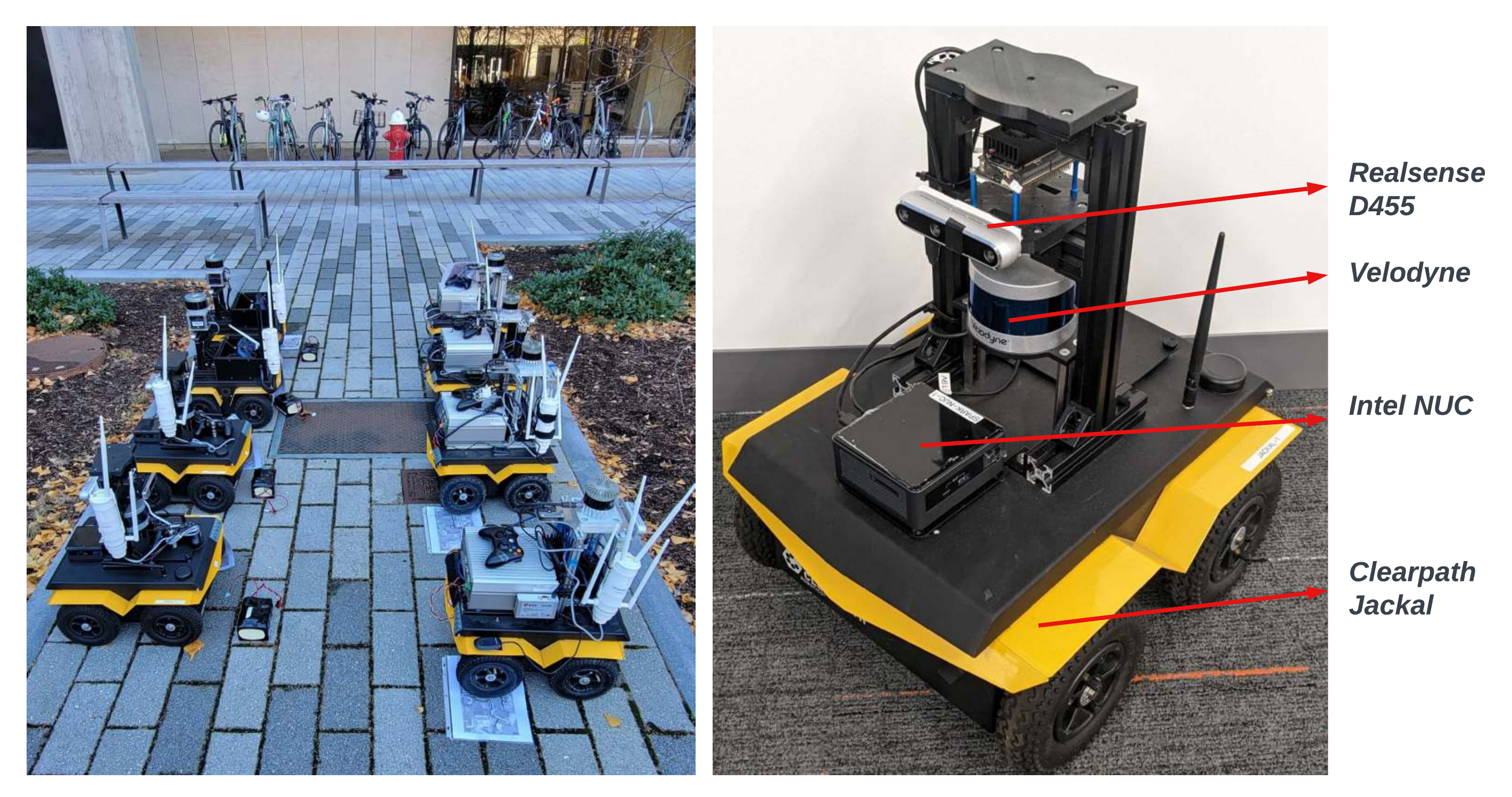}
    \caption{The Jackal robots used for the experiments.}\label{fig:jackals}
    \vspace{-3mm}
\end{figure}

\begin{figure}[!t]
    \centering
    \includegraphics[trim={0 0 0 0},clip, width=1.0\columnwidth]{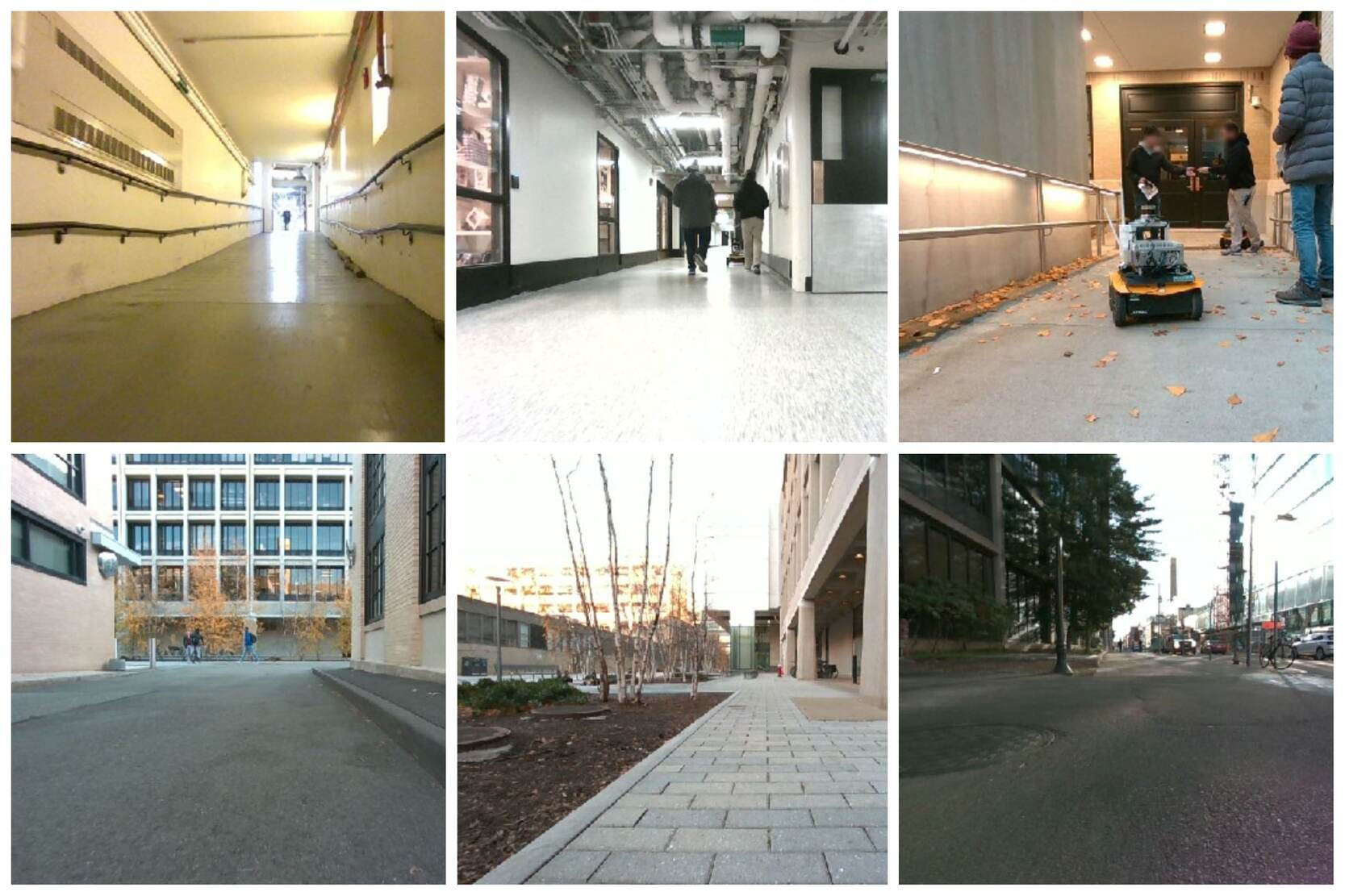}
    \caption{Snapshots from the \Hybrid dataset.}\label{fig:dataset_photos}
    \vspace{-6mm}
\end{figure}

We conducted live experiments on the MIT campus with 
a fleet of Clearpath Jackal rovers
equipped with a Realsense D455 RGB-D Camera and an Ouster or Velodyne 3D lidar (Fig.~\ref{fig:jackals}).
Each robot has an Intel NUC computer with an Intel i7 4.70 GHz processor,
and communicates with the other robots via a 2.4 GHz wireless network.
Each robot was tele-operated by a human operator.
During the experiment, there were often humans and vehicles moving around the robots,
and the robots traverse through a variety of both indoor and outdoor environments (Fig.~\ref{fig:dataset_photos}).

\begin{figure}[t]
	\centering
	\subfloat[\Outdoor]{%
		\includegraphics[trim=0 0 0 0, clip, width=0.48\columnwidth]
		{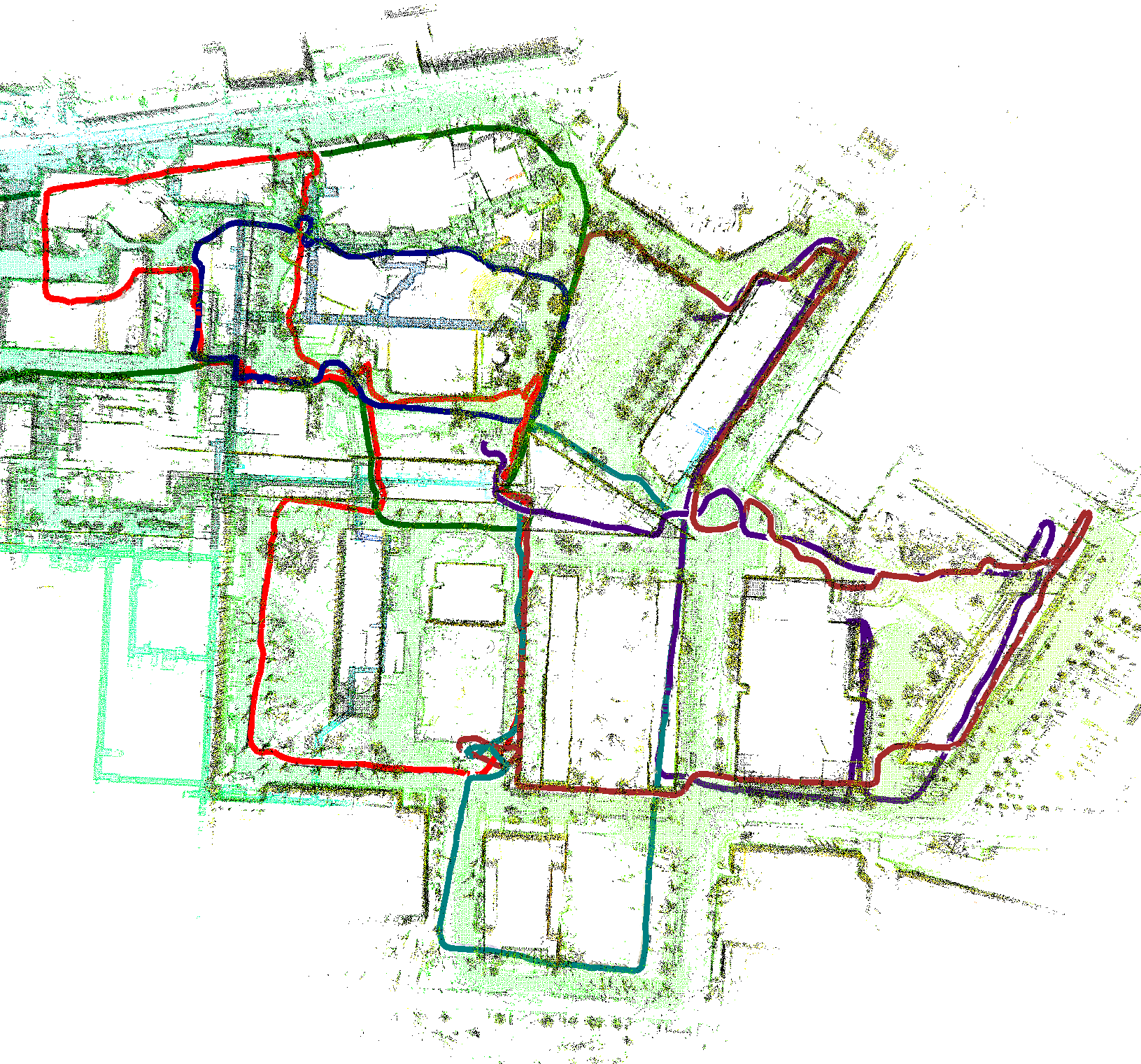}
		\label{fig:campus_outdoor}
	} ~
	\subfloat[\Tunnels]{%
		\includegraphics[trim=0 0 0 0, clip, width=0.48\columnwidth]
		{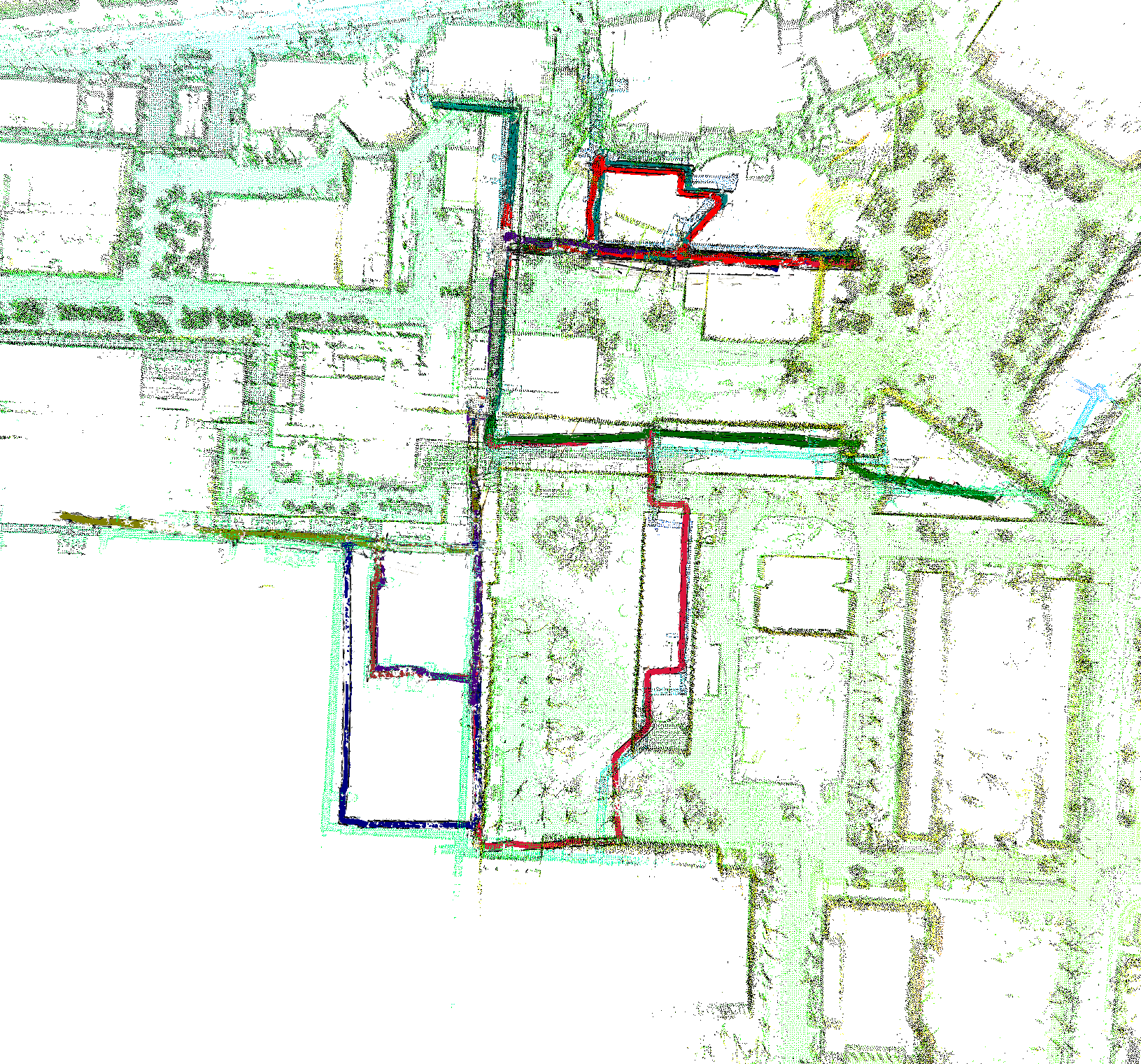}
		\label{fig:campus_tunnels}
	}
	\caption{\small Estimated trajectories, estimated meshes, and reference point-cloud for the \Outdoor (6 robots, $6044$m)
		and \Tunnels (8 robots, $6753$m) datasets.}
	\label{fig:datasets}
	\vspace{-4mm}
\end{figure}

We compiled three real-world datasets based on the data recorded. 
The first dataset is the \Outdoor dataset (Fig.~\ref{fig:campus_outdoor}),
which consists of six Jackal rovers traversing a total of 6044 meters over a duration of around 20 minutes on the MIT campus. 
Except for one of the robots traversing a crowded building, most of the dataset consists of outdoor campus and urban scenes.
The second dataset is the \Tunnels dataset (Fig.~\ref{fig:campus_tunnels}),
which consists of eight Jackal rovers traversing a total of 6753 meters over a duration of almost 30 minutes in the tunnels underneath the university campus.
The trajectories are entirely indoors and consist mostly of long homogeneous tunnels and corridors. 
The last dataset is the \Hybrid dataset (Fig.~\ref{fig:cover}),
which consists of eight Jackal rovers traversing a total of 7785 meters over a duration of almost 30 minutes both on and below the university campus.
This dataset is also multi-level, with some trajectories crossing each other on different levels.

The reference trajectory we use as ground truth for evaluation is generated by first building a reference point-cloud map as shown in the background of Fig.~\ref{fig:cover}
using lidar-based SLAM~\cite{Chang22ral-LAMP2,Reinke22ral-LOCUS2},
with additional total station measurements for the indoor portions of the datasets
and differential GPS measurements throughout the outdoor areas in and around the university campus.
Once we have the reference point-cloud map,
we obtain the reference trajectories using LOCUS 2.0~\cite{Reinke22ral-LOCUS2} by matching the lidar point-cloud for each robot at each time-step
against the reference point-cloud map. 

\section{Experiments}
\label{sec:experiments}

In this section, we present a quantitative analysis on the accuracy and resilience of \kimeraMulti.
We perform the experiments by replaying our datasets in real-time on robots connected via a wireless network (see Section~\ref{sec:experiments_setup} for details),
which allows us to evaluate different communication scenarios and obtain comprehensive statistics.
Results from live field tests are discussed in Section~\ref{sec:discussion}.

\subsection{Experimental Setup}
\label{sec:experiments_setup}

The experiments in this section are obtained by playing back the datasets on robots in real-time.
All agents communicate with each other via a 2.4~GHz wireless network.
On top of the real communication, we simulate different types of disconnections (discussed in the next paragraph)
by controlling connectivity in the remote topic manager module (Section~\ref{sec:system}).
Apart from the simulated communication disruptions,
other aspects of the experiments such as available onboard compute and the rate at which the input is received by \kimeraMulti
are equivalent to the live experiments.

{\bf Communication Scenarios.}
We evaluate our system under four simulated communication scenarios (implemented in the remote topic manager):\footnote{The simulation applies to all robots, and optionally to the base station (placed at the origin) when running the centralized baseline.
	For the purpose of computing the trajectory error,
	for each dataset we start the simulation after at least two robots are initialized in the global frame.}
(i) \scenario{Full}: all nodes are connected at all times;
(ii) \scenario{Random}: each node is disconnected randomly three times during the mission, and each disconnection lasts 90 seconds.
This scenario tests the resilience of our system to situations in which some robots experience temporary failures and go offline for some time;
(iii) \scenario{Distance}: connection is established between nodes within a distance threshold.
Note that under this setup, there can be multiple clusters of connected nodes at any given time.
In our experiments, we set the distance threshold to $75$m, which we observe to create interesting and challenging network topologies 
on all datasets.
We assume robots in each connected component can help route information to others, so that all robots in the same cluster can communicate with each other;
(iv) \scenario{Base}: nodes can communicate only within a certain radius of the base station ($75$m and same as \scenario{Distance}).
This setting is the closest to the communication setup observed in our live experiments.

{\bf Centralized Baseline.}
We implement a centralized baseline system that detects all inter-robot loop closures at a base station.
The back-end subscribes to the pose graphs and the loop closures from the robots,
and optimizes the full problem using GNC with Levenberg-Marquardt as implemented with GTSAM~\cite{gtsam}.
The loop closure parameters, GNC parameters, along with other related parameters like BoW vector downsampling and 
pose graph coarsening are consistent with the distributed setup.
Additionally, details such as the initial reference frame alignment were implemented to be congruent to its distributed counterpart.
For our experiments, the base station is ran on a laptop with an Intel i9 2.40~GHz processor.

\subsection{Real-time Evaluation Under Unreliable Communication}
\label{sec:experiments_monte_carlo}
In the following, 
we first discuss the key statistics 
from the front-end, the back-end, and the communication system.
Then, 
we discuss in depth the performance for each of the data sequences under each communication scenario.

\begin{table*}[t]
\setlength{\tabcolsep}{2pt}
\centering
\caption{\small Summary of Front-end and Back-end Statistics for \kimeraMulti (Dist) and centralized baseline (Cent).}\label{tab:summary}
\begin{tabular}{cc |ccc|ccccccc}
\toprule
& & \multicolumn{3}{c}{Front-end} & \multicolumn{7}{|c}{Back-end} \\
\midrule
& & \multicolumn{3}{c}{Loop Closures} & \multicolumn{1}{|c}{Size} & \multicolumn{2}{c}{Iterations} & \multicolumn{2}{c}{Optimization Time} & \multicolumn{2}{c}{Metric Error} \\
\midrule
& & \# Matches & \# Loops & \% Inliers & \# Poses & \# Max & \# Median & Max(s) & Avg(s) & ATE(m) & AME(m) \\
\midrule
\multirow{3}{*}{Dist}
& Tunnels & 12397 $\pm$ 74& 10620 $\pm$ 70& 20.8 $\pm$ 0.3 & 2832 $\pm$ 6& 1083 $\pm$ 23& 571 $\pm$ 68& 285.55 $\pm$ 31.74& 139.52 $\pm$ 15.52& 4.48 $\pm$ 0.29& 2.62 $\pm$ 0.07\\
& Hybrid & 2999 $\pm$ 136& 1252 $\pm$ 30& 26.9 $\pm$ 0.2 & 3201 $\pm$ 6& 1190 $\pm$ 62& 698 $\pm$ 133& 336.98 $\pm$ 26.25& 174.05 $\pm$ 14.97& 7.71 $\pm$ 0.78& 1.9 $\pm$ 0.26\\
& Outdoor & 714 $\pm$ 19& 172 $\pm$ 10& 36.5 $\pm$ 2.3 & 2383 $\pm$ 1& 735 $\pm$ 327& 344 $\pm$ 31& 196.9 $\pm$ 43.11& 100.92 $\pm$ 11.46& 12.4 $\pm$ 1.92& 4.83 $\pm$ 1.35\\
\midrule
\multirow{3}{*}{Cent}
& Tunnels & 12432 $\pm$ 70& 8247 $\pm$ 188& 19.8 $\pm$ 0.4 & 3053 $\pm$ 4& - & - & 11.51 $\pm$ 1.05& 3.24 $\pm$ 0.28& 4.38 $\pm$ 0.21& 2.58 $\pm$ 0.12\\
& Hybrid & 3052 $\pm$ 45& 1302 $\pm$ 18& 24.7 $\pm$ 0.5 & 3508 $\pm$ 1& - & - & 11.44 $\pm$ 0.21& 1.65 $\pm$ 0.16& 5.83 $\pm$ 0.16& 1.58 $\pm$ 0.05\\
& Outdoor & 732 $\pm$ 6& 182 $\pm$ 5& 37.4 $\pm$ 1.0 & 2647 $\pm$ 0& - & - & 11.49 $\pm$ 2.94& 0.55 $\pm$ 0.12& 9.38 $\pm$ 0.31& 4.99 $\pm$ 0.48\\
\midrule
\end{tabular}
\vspace{-5mm}
\end{table*}
{\bf Front-end Statistics.}
We present a summary of front-end statistics in Table~\ref{tab:summary}.
The number of BoW matches, detected loop closures, and the percentage of inlier loop closures
are summarized over 3 trials under the \scenario{Full} communication scenario.
A loop closure is classified as an inlier if the corresponding relative transformation 
is within 10 cm in translation and 10 degrees in rotation compared to the ground truth.
We only present the summary from the \scenario{Full} communication scenario.
Results under other scenarios are quantitatively similar.

\begin{figure}[!t]
	\centering
	\includegraphics[trim={0 0 0 0},clip, width=0.6\columnwidth]
	{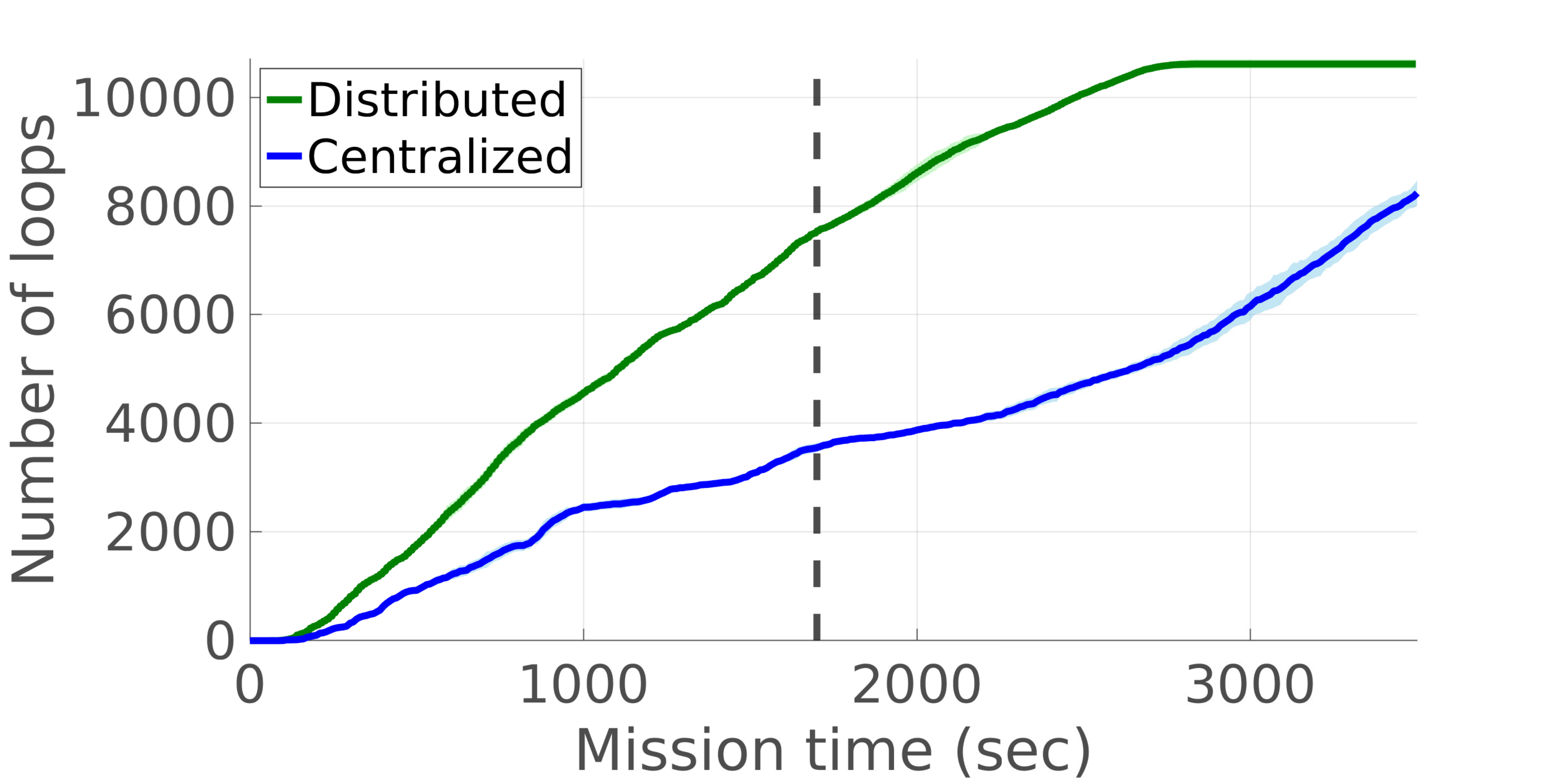}
	\caption{\small Number of detected loop closures on the \Tunnels dataset under the \scenario{Full} communication scenario. Vertical dashed line denotes the time when all robots finish exploration.}\label{fig:lcd_tunnels_full}
	\vspace{-3mm}
\end{figure}

Significantly more loop closures are detected on the indoor \Tunnels dataset
as there are smaller viewpoint variations in narrow corridors compared to outdoor areas that are generally more spacious.
In general, 
the distributed system and the centralized baseline detect similar number of loop closures with similar inlier percentage.
On the \Tunnels dataset, however, 
the distributed front-end detects significantly more loop closures than the centralized baseline.
This is because we terminate the experiment at $3500$ seconds (more than twice the duration of the actual dataset),
and the centralized system has not yet finished processing all the loop closures; see Fig.~\ref{fig:lcd_tunnels_full}.
Lastly, we note that there is a significant number of outlier loop closures detected across all datasets,
which highlights the importance of implementing an outlier-robust back-end.

\begin{table}[t]
\setlength{\tabcolsep}{2pt}
\centering
\caption{\small Summary of Communication Statistics.}\label{tab:comms_summary}
\begin{tabular}{cccccc}
\toprule
& \multicolumn{2}{c}{Delay} & \multicolumn{2}{c}{Bandwidth} & Packet Drop \\
\midrule
& Max(s) & Avg(s)& Max(MB/s) & Avg(MB/s) & \% Drop \\
\midrule
Tunnels & 7.71 $\pm$ 1.65& 0.10 $\pm$ 9e-3& 2.10 $\pm$ 0.02& 0.55 $\pm$ 0.04& 45.55 $\pm$ 7.89\\
Hybrid & 4.17 $\pm$ 0.98& 0.05 $\pm$ 3e-3& 0.41 $\pm$ 0.02& 0.12 $\pm$ 4e-3& 21.50 $\pm$ 5.41\\
Outdoor & 1.53 $\pm$ 0.36& 0.05 $\pm$ 3e-3& 0.10 $\pm$ 0.01& 0.01 $\pm$ 2e-3& 20.28 $\pm$ 5.72\\
\midrule
\end{tabular}
\vspace{-8mm}
\end{table}

{\bf Back-end Statistics.}
Table~\ref{tab:summary} also presents the statistics from the distributed and centralized back-end.
In particular, the following metrics are reported: 
(1) the number of poses in the PGO problem, 
(2) the maximum and median number of iterations taken by the distributed back-end,
(3) the maximum and average optimization time,
(4) the final absolute trajectory error (ATE), 
and (5) the final average map error (AME).
Among these metrics, the ATE is defined as the root-mean-square error between the estimated and reference trajectories,
and the AME is defined as the average mesh-vertex-to-point distance between the estimated mesh and the reference point-cloud map.
All metrics are computed by summarizing the results over 3 trials taken from the 
\scenario{Full} communication scenario.

For all three datasets, the distributed back-end achieves comparable but less accurate results compared to the centralized back-end,
due to the fact that the distributed solver only solves GNC approximately \cite{Tian21tro-KimeraMulti}.
The distributed back-end also has a much higher optimization time, mainly because of
the large number of iterations required by distributed PGO and the communication overhead
accumulated across iterations.
{We remark that Table~\ref{tab:summary} compares the centralized and distributed system under \scenario{Full} communication; later in this section we discuss how different communication scenarios create a more interesting trade-off between centralized and distributed architectures.}

{\bf Communication Statistics.}
Table~\ref{tab:comms_summary} presents the communication profile of \kimeraMulti.
The results are averaged over all pairs of robots from 3 trials under the \scenario{Full} communication scenario.
During the experiments, 
there are sometimes delays of up to 7 seconds and a substantial amount of dropped packets (up to $45\%$),
which demonstrates the challenges of using real-world wireless communication.
The other three communication scenarios introduce even more challenges in the form of long periods of disconnections.
Overall, our results demonstrate that
\kimeraMulti is resilient to imperfect communication with large delays and unreliable packet delivery.

\begin{figure*}[ht]
	\centering
	\subfloat[\Tunnels (Full)]{%
		\includegraphics[trim=40 0 70 30, clip, width=0.24\textwidth]
		{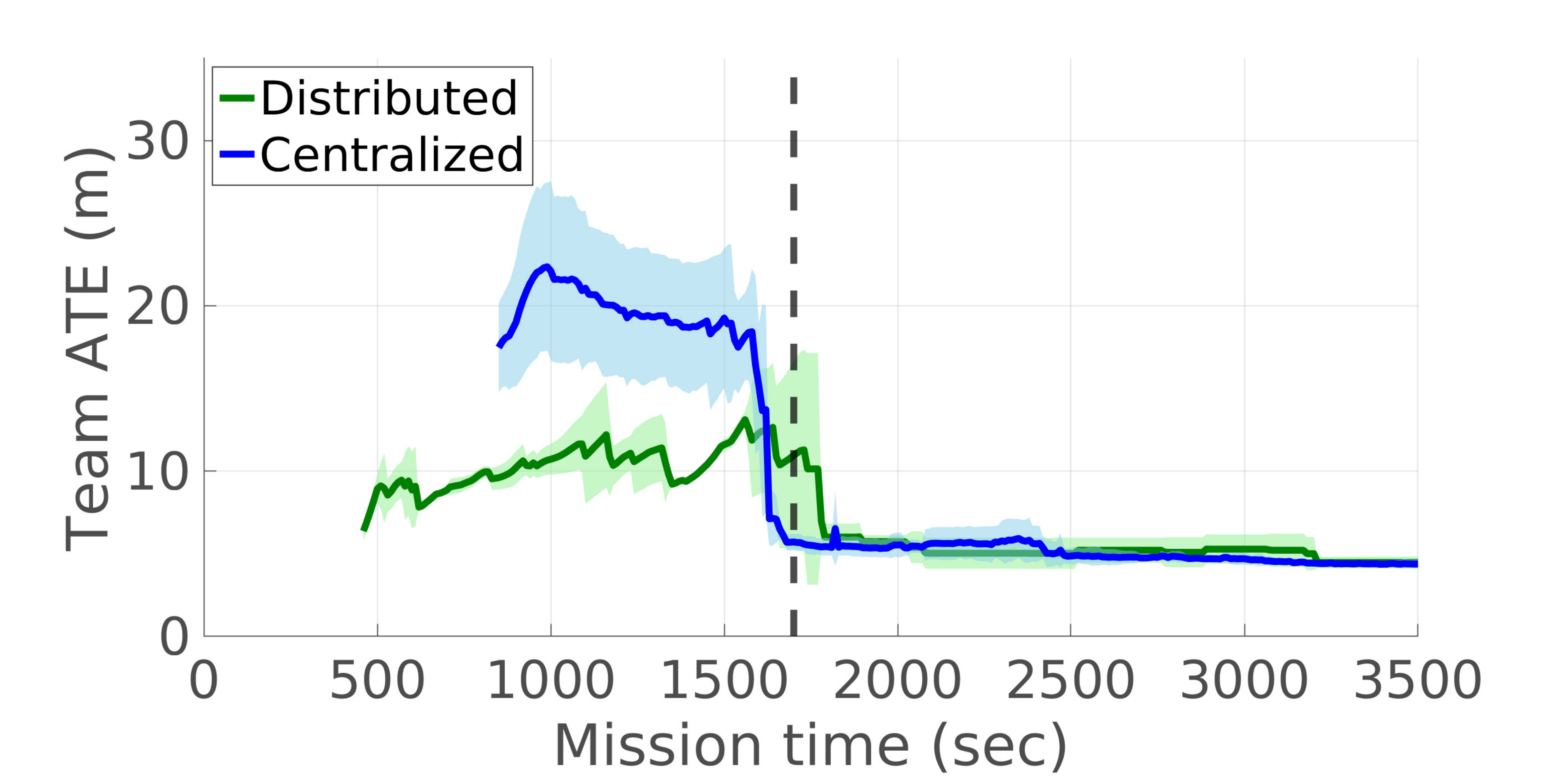}
		\label{fig:ate_tunnels_full}
	} ~
	\subfloat[\Tunnels (Random)]{%
		\includegraphics[trim=40 0 70 30, clip, width=0.24\textwidth]
		{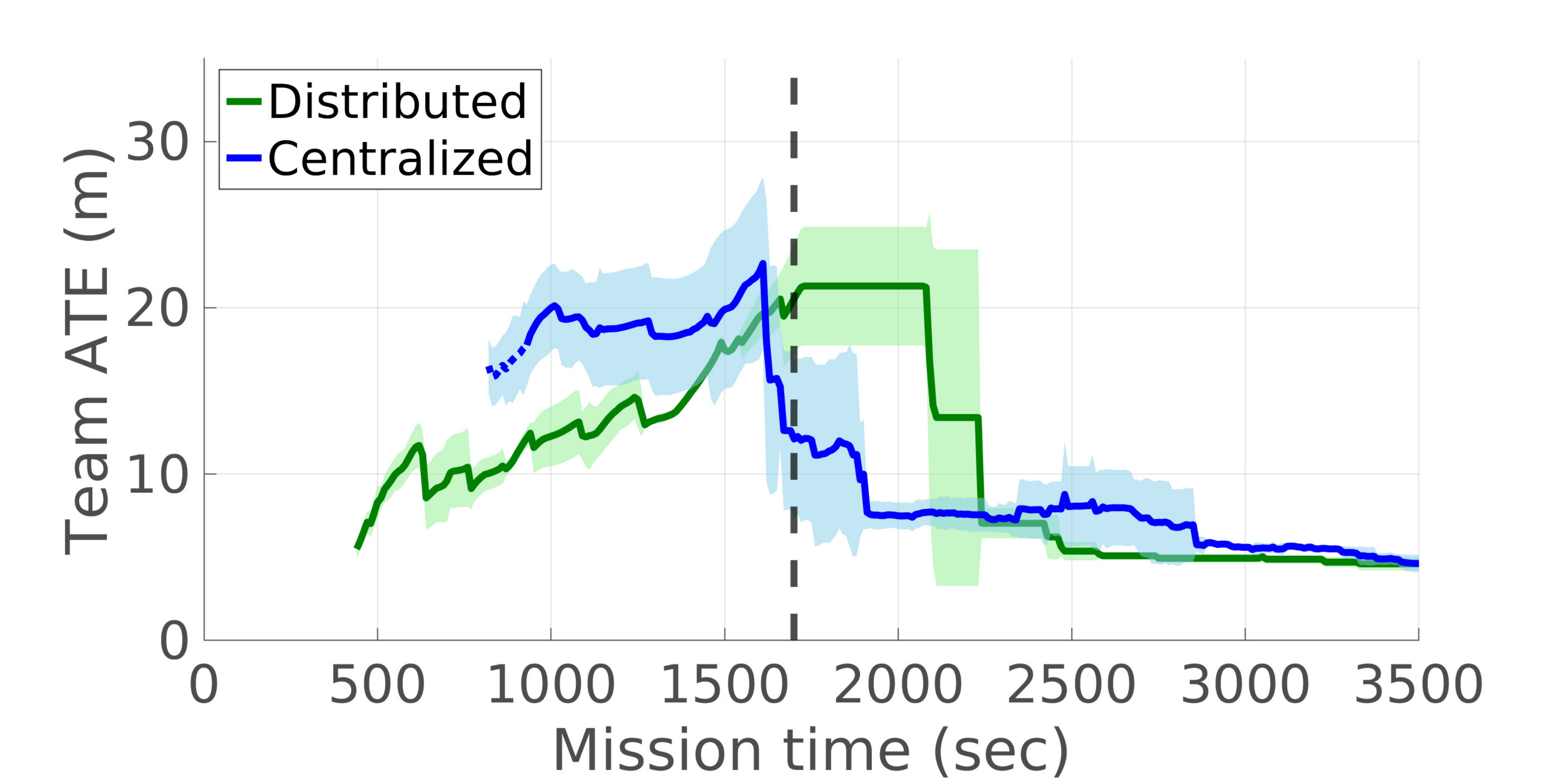}
		\label{fig:ate_tunnels_random}
	}~
	\subfloat[\Tunnels (Distance)]{%
		\includegraphics[trim=40 0 70 30, clip, width=0.24\textwidth]
		{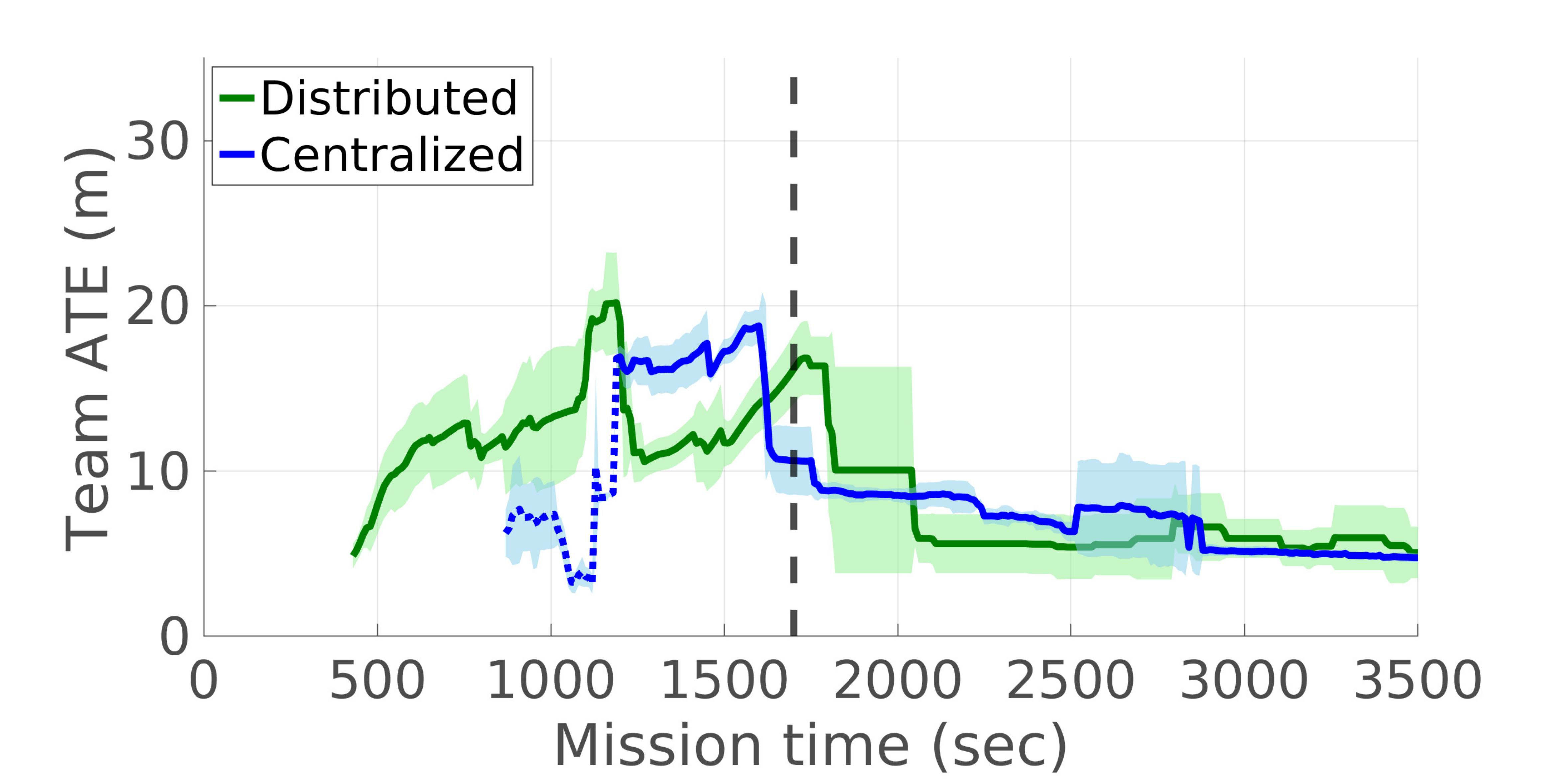}
		\label{fig:ate_tunnels_distance}
	}
	~
	\subfloat[\Tunnels (Base)]{%
		\includegraphics[trim=40 0 70 30, clip, width=0.24\textwidth]
		{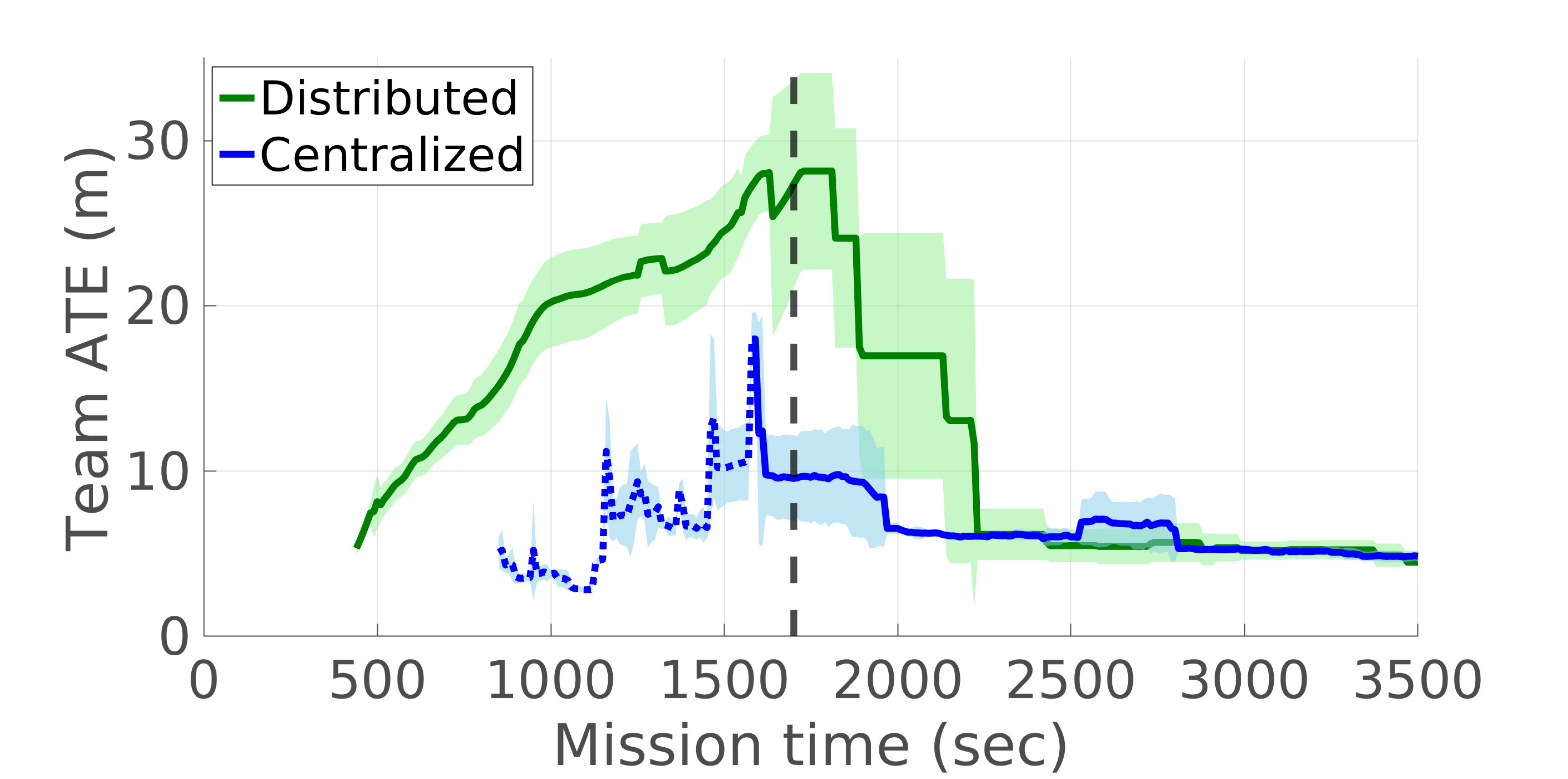}
		\label{fig:ate_tunnels_base}
	}
	\\
	\subfloat[\Hybrid (Full)]{%
		\includegraphics[trim=40 0 70 30, clip, width=0.24\textwidth]
		{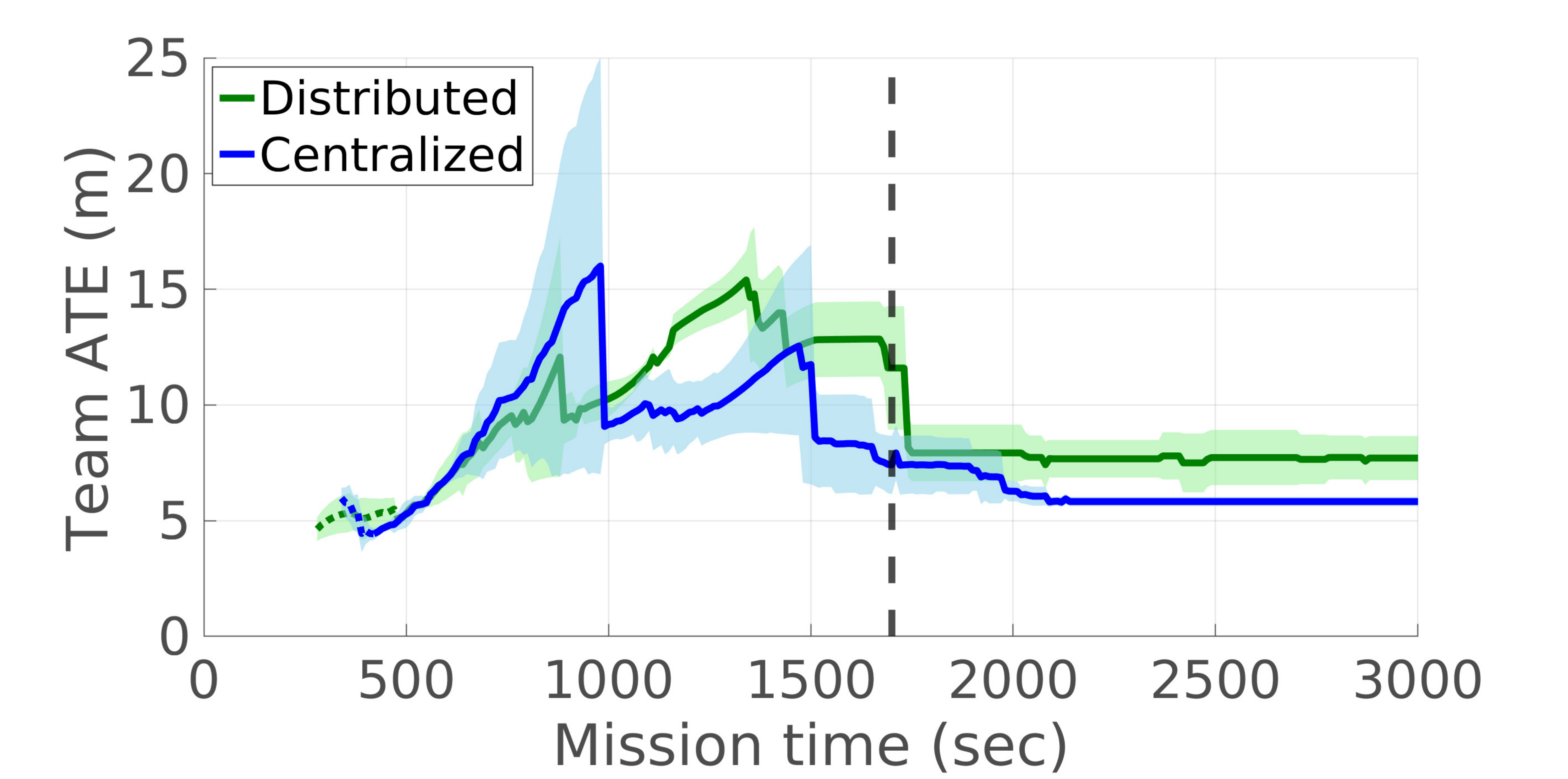}
		\label{fig:ate_hybrid_full}
	} ~
	\subfloat[\Hybrid (Random)]{%
		\includegraphics[trim=40 0 70 30, clip, width=0.24\textwidth]
		{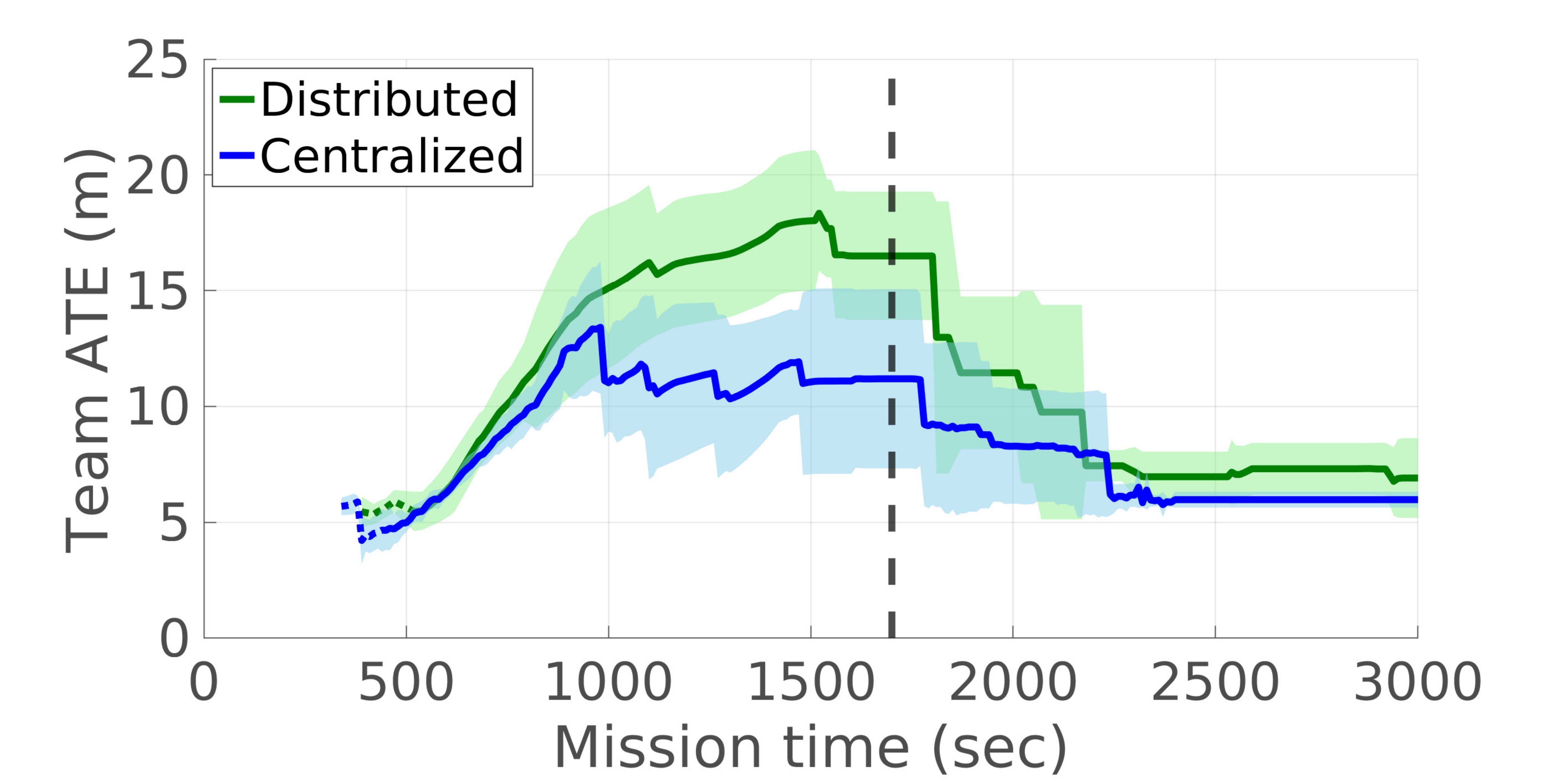}
		\label{fig:ate_hybrid_random}
	}~
	\subfloat[\Hybrid (Distance)]{%
		\includegraphics[trim=40 0 70 30, clip, width=0.24\textwidth]
		{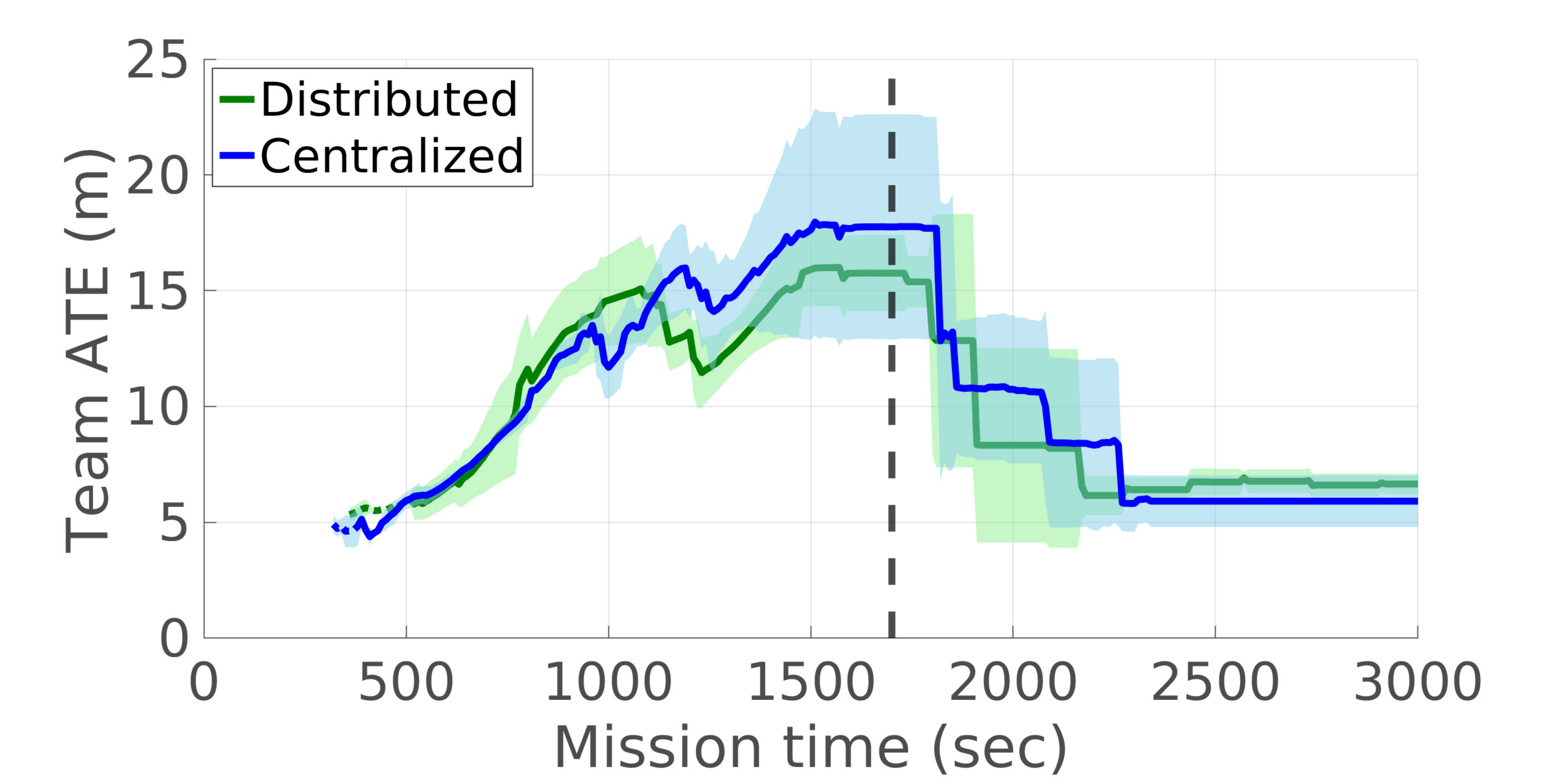}
		\label{fig:ate_hybrid_distance}
	}
	~
	\subfloat[\Hybrid (Base)]{%
		\includegraphics[trim=40 0 70 30, clip, width=0.24\textwidth]
		{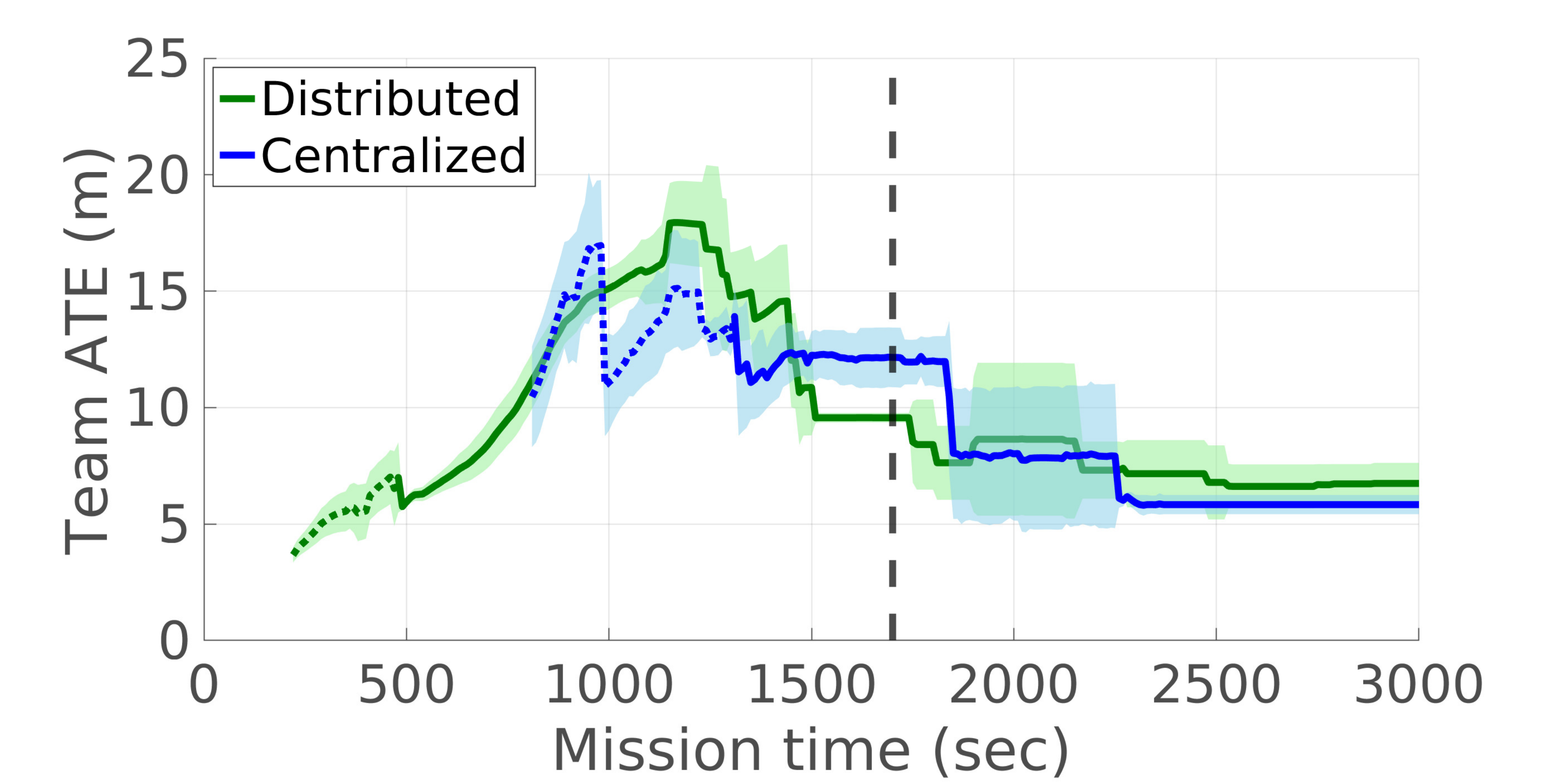}
		\label{fig:ate_hybrid_base}
	}
	\\
	\subfloat[\Outdoor (Full)]{%
		\includegraphics[trim=40 0 70 30, clip, width=0.24\textwidth]
		{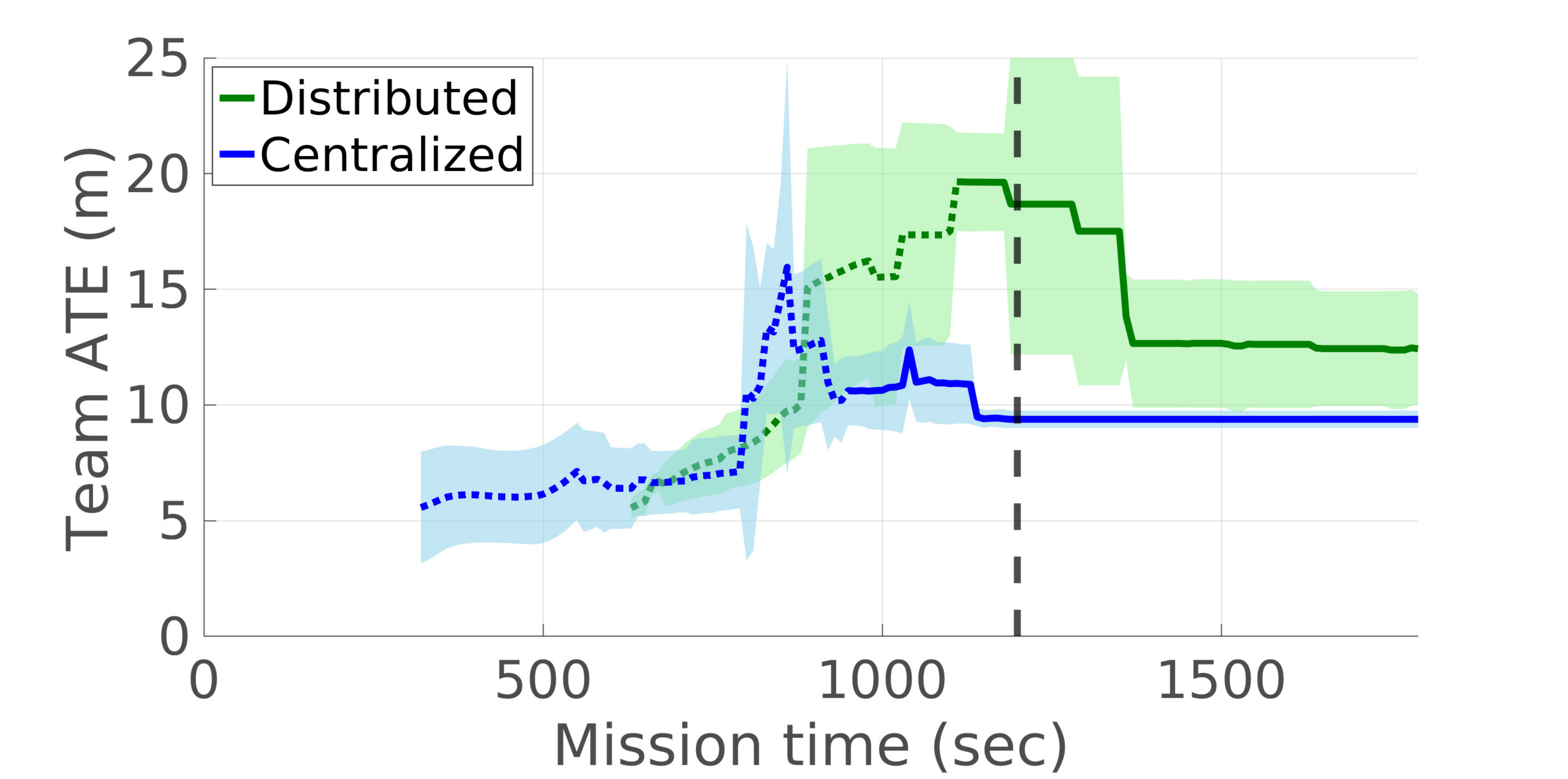}
		\label{fig:ate_outdoor_full}
	} ~
	\subfloat[\Outdoor (Random)]{%
		\includegraphics[trim=40 0 70 30, clip, width=0.24\textwidth]
		{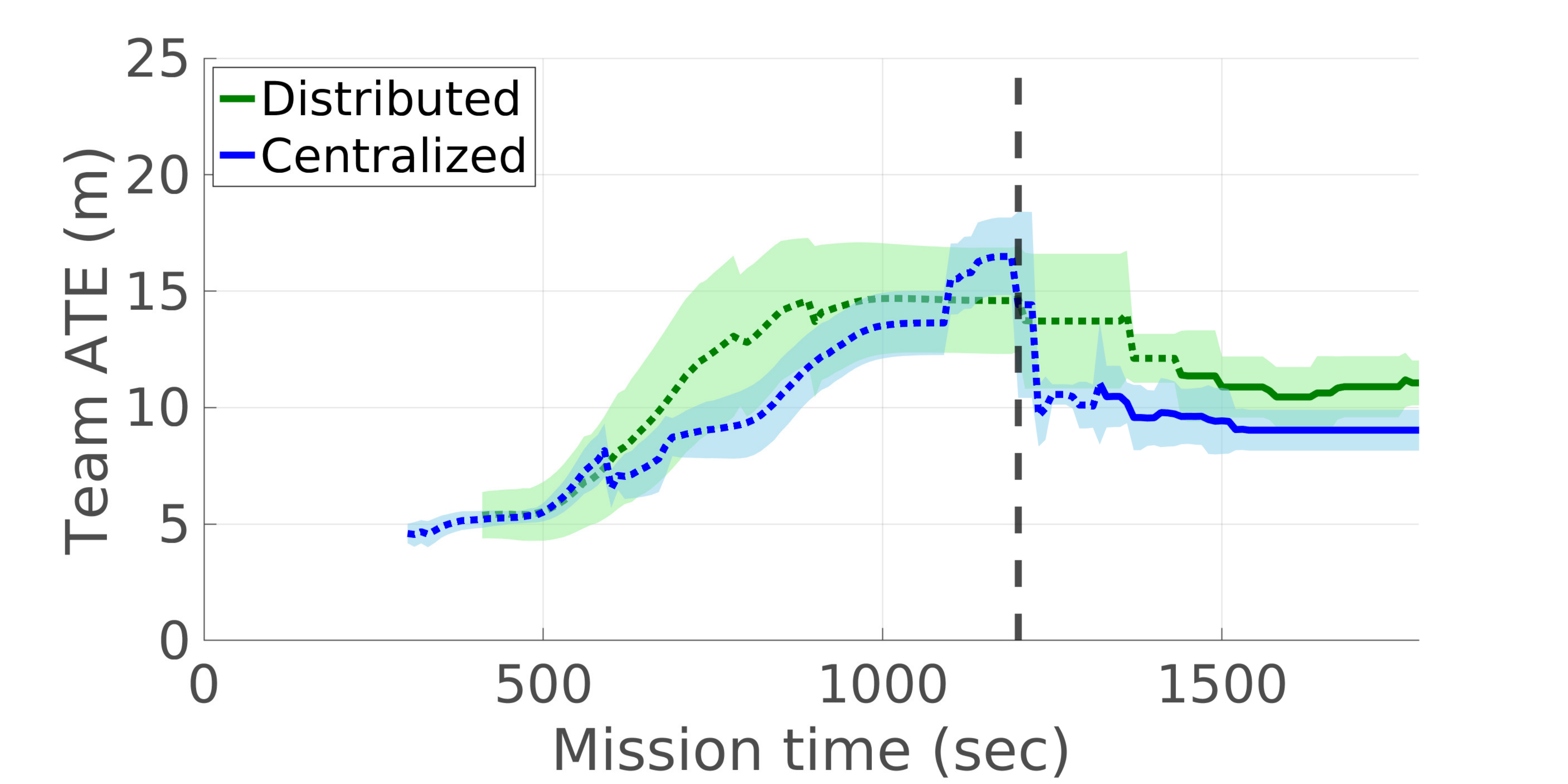}
		\label{fig:ate_outdoor_random}
	}~
	\subfloat[\Outdoor (Distance)]{%
		\includegraphics[trim=40 0 70 30, clip, width=0.24\textwidth]
		{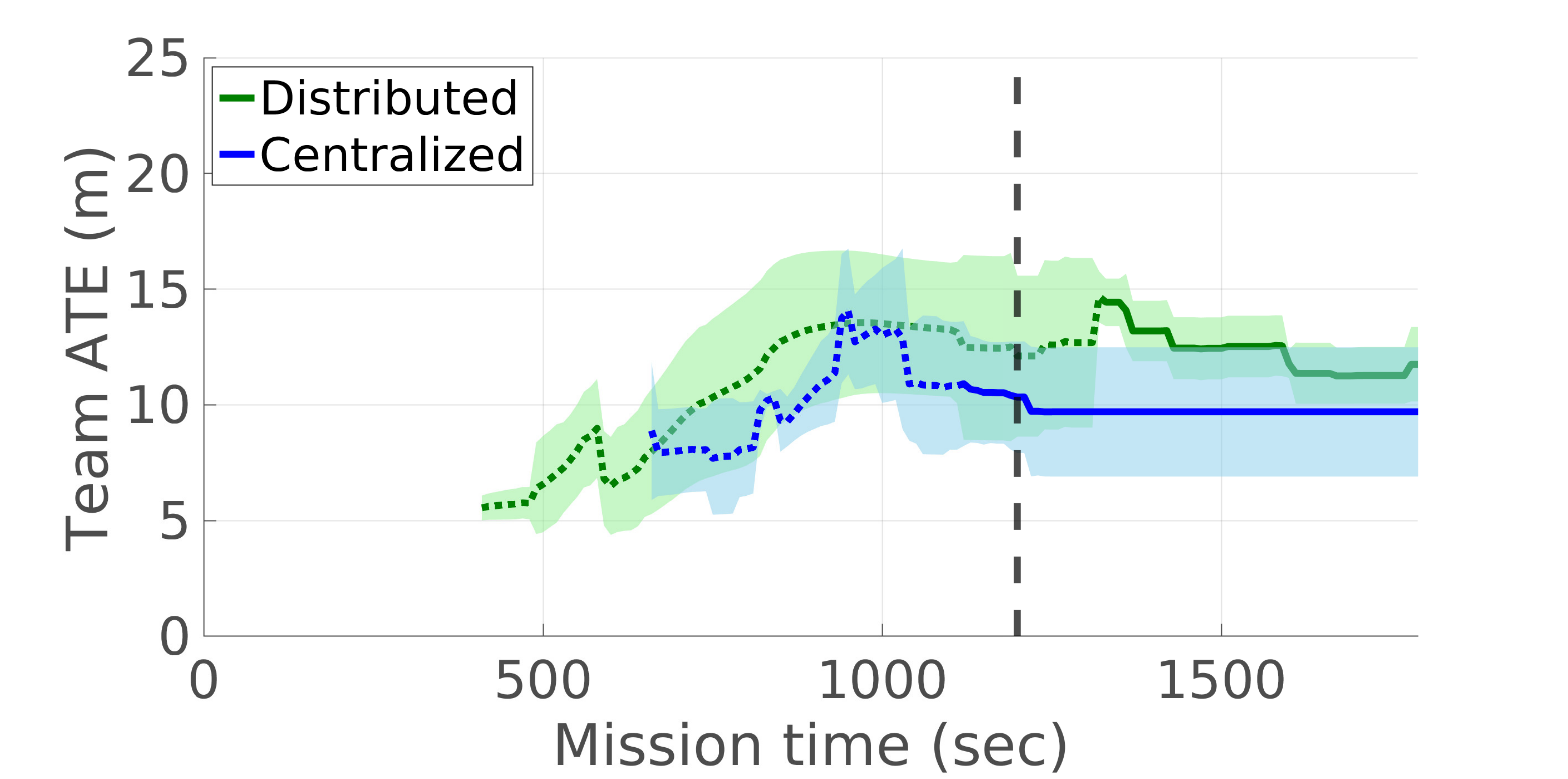}
		\label{fig:ate_outdoor_distance}
	}
	~
	\subfloat[\Outdoor (Base)]{%
		\includegraphics[trim=40 0 70 30, clip, width=0.24\textwidth]
		{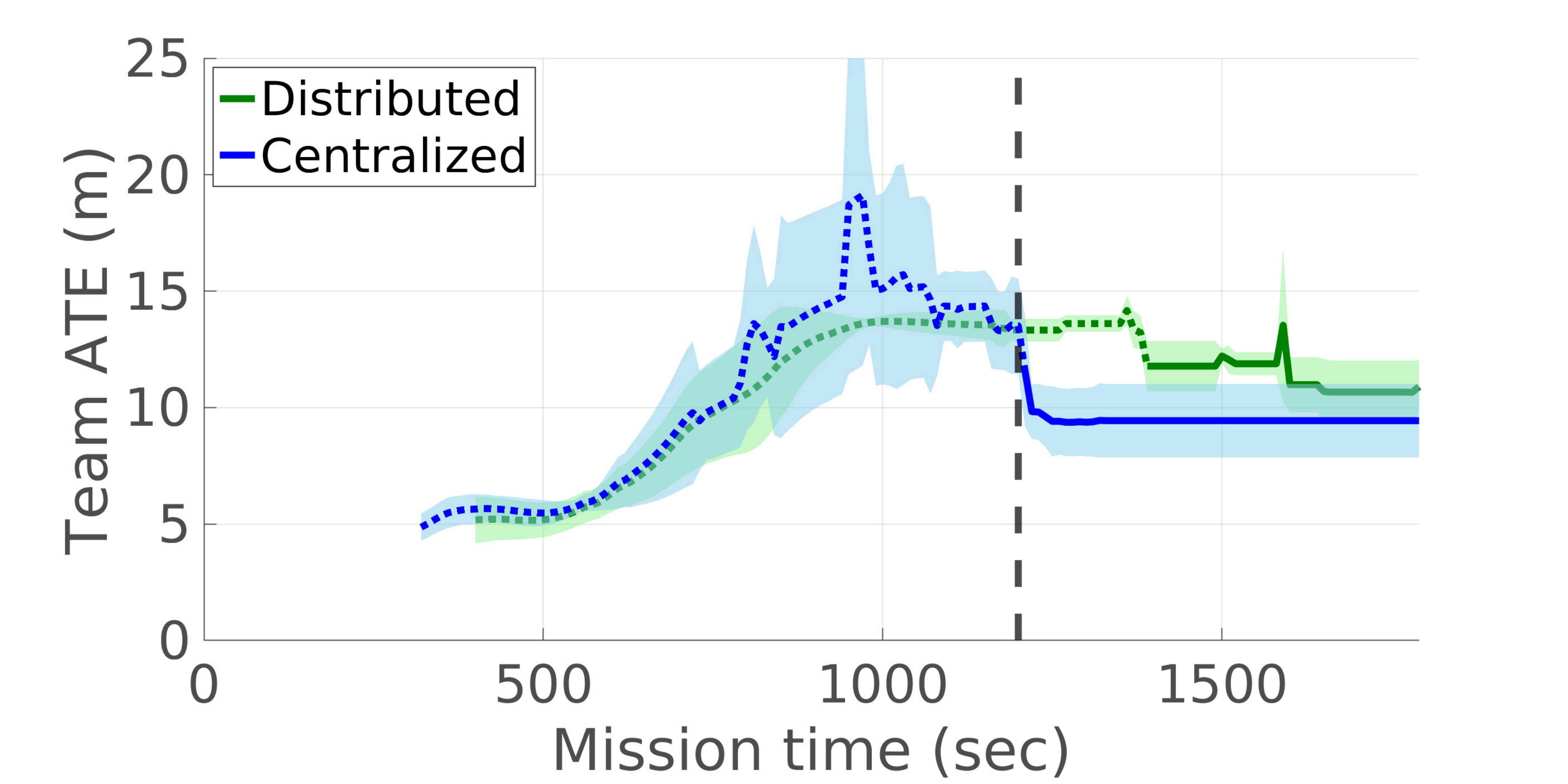}
		\label{fig:ate_outdoor_base}
	}
	\caption{\small ATE evaluation under different communication scenarios.
		Each row shows results on a single dataset, and each column corresponds to a communication scenario.
		Dashed line indicates that only a subset of robots is estimated,
		whereas solid line indicates that all robots are estimated. }
	\label{fig:ate_eval}
	\vspace{-6mm}
\end{figure*}

{\bf Evolution of ATE.}
In Fig.~\ref{fig:ate_eval}, we visualize the evolution of the ATE on the \Tunnels, \Hybrid, and \Outdoor datasets
under all four communication scenarios.
Each plot includes the ATE of both the distributed system and the centralized baseline,
where the line represents the average ATE over 3 trials and the shaded area shows one standard deviation.
The dashed portion of each line indicates that only a subset of robots is initialized in the global frame at that time,
and consequently the ATE is only computed over that subset of robots.
The line turns solid as soon as all robots are initialized in the global frame,
at which point the ATE is computed over all robots.
Finally, the vertical dashed line in each plot denotes the end of the mission (\ie all robots finish mapping.)

The \Tunnels dataset represents a scenario with many loop closures (see Table~\ref{tab:summary}).
For the \scenario{Full} communication scenario (Fig.~\ref{fig:ate_tunnels_full}),
the distributed system is able to detect enough loop closures to initialize all robots in the global frame sooner
and also achieves a lower ATE in the earlier portions of the data sequence. 
This is mostly due to the faster processing of the distributed front-end, which parallelizes loop closure detection across robots.
However, the distributed back-end is slower than the centralized back-end, and the centralized system converges quickly
after most of the loop closures are detected. 
At steady state, the performance of the centralized and distributed systems are similar.
For the \scenario{Random} scenario (Fig.~\ref{fig:ate_tunnels_random}), 
we observe that random disconnects in the communication leads to a delayed decrease in ATE for both
the distributed and centralized systems.
However, the effect is more significant for the distributed system as the distributed back-end is interrupted by disconnections.
The centralized system behaves similarly for the \scenario{Distance} and \scenario{Base} scenarios,
but the distributed system performs better under the \scenario{Distance} scenario (Fig.~\ref{fig:ate_tunnels_distance})
compared to the \scenario{Base} scenario (Fig.~\ref{fig:ate_tunnels_base}).
This is because the distributed system
is able to both detect loop closures and perform distributed PGO while being disconnected from the base station
(as shown by the drop in ATE around 1300 seconds in Fig.~\ref{fig:ate_tunnels_distance}).

In the \Hybrid dataset,
we again see the advantage of the distributed front-end as the distributed system stabilizes faster
than the centralized baseline for the \scenario{Full} scenario (Fig.~\ref{fig:ate_hybrid_full}).
For the \scenario{Random} scenario (Fig.~\ref{fig:ate_hybrid_random}), the overall ATE is increased for the distributed system due to delays caused by random disconnects.
The distributed system under the \scenario{Distance} (Fig.~\ref{fig:ate_hybrid_distance}) and \scenario{Base} (Fig.~\ref{fig:ate_hybrid_base}) scenarios
has similar performance to the centralized system throughout the mission,
but the distributed system under the \scenario{Distance} scenario 
is able to maintain lower and more stable ATE
between 1200 to 1800 seconds
due to its flexibility to detect loop closures and perform PGO while being disconnected from the base station.

On the \Outdoor dataset, the distributed front-end has little advantage
due to the much smaller number of loop closures (see Table~\ref{tab:summary}).
This dataset is also challenging for the distributed back-end,
as the sparse loop closures make the underlying optimization problem poorly conditioned and
the distributed solver struggles to reach a good precision solution.

{In summary, the distributed front-end is able to speed up loop closure detection by distributing the computational load across the robots; on the other hand, the distributed back-end requires substantially more time to converge compared to its centralized counterpart, but in some cases, it provides extra flexibility by enabling clusters of robots to optimize their trajectories while being away from the base station. This translates into reduced ATE in some portions of the trajectories (\eg Fig.~\ref{fig:ate_tunnels_full}, \ref{fig:ate_tunnels_random}, and \ref{fig:ate_hybrid_distance}).}

\subsection{Parameter Sensitivity}
\label{sec:experiments_parameters}

In this section, we present a detailed analysis of three key parameters that are observed to directly impact the accuracy of our system.
These include two parameters related to our distributed solver:
the gradient norm threshold $\epsilon_{g}$ that controls the precision when robots solve their local optimization problems, 
and the relative change threshold $\epsilon_{\text{rel}}$ that determines when to terminate distributed optimization (\Cref{sec:system}).
Intuitively, setting smaller values for these thresholds enable distributed PGO to obtain more accurate results at the expense of increasing iterations and runtime.
The last parameter we analyze is the distance threshold $d$ used to coarsen the pose graph.
Recall from \Cref{sec:system} that our system aggregates nearby pose graph nodes within this distance threshold.
Thus, larger values of $d$ yield a smaller and coarser pose graph.
As we vary each parameter, we fix the remaining parameters to their nominal values ($\epsilon_{g}=0.1$, $\epsilon_{\text{rel}}=0.2$m, and $d=2$m).

\begin{figure}[t]
	\centering
	\subfloat[$\epsilon_{g}$]{%
		\includegraphics[trim=20 0 100 0, clip, width=0.14\textwidth]
		{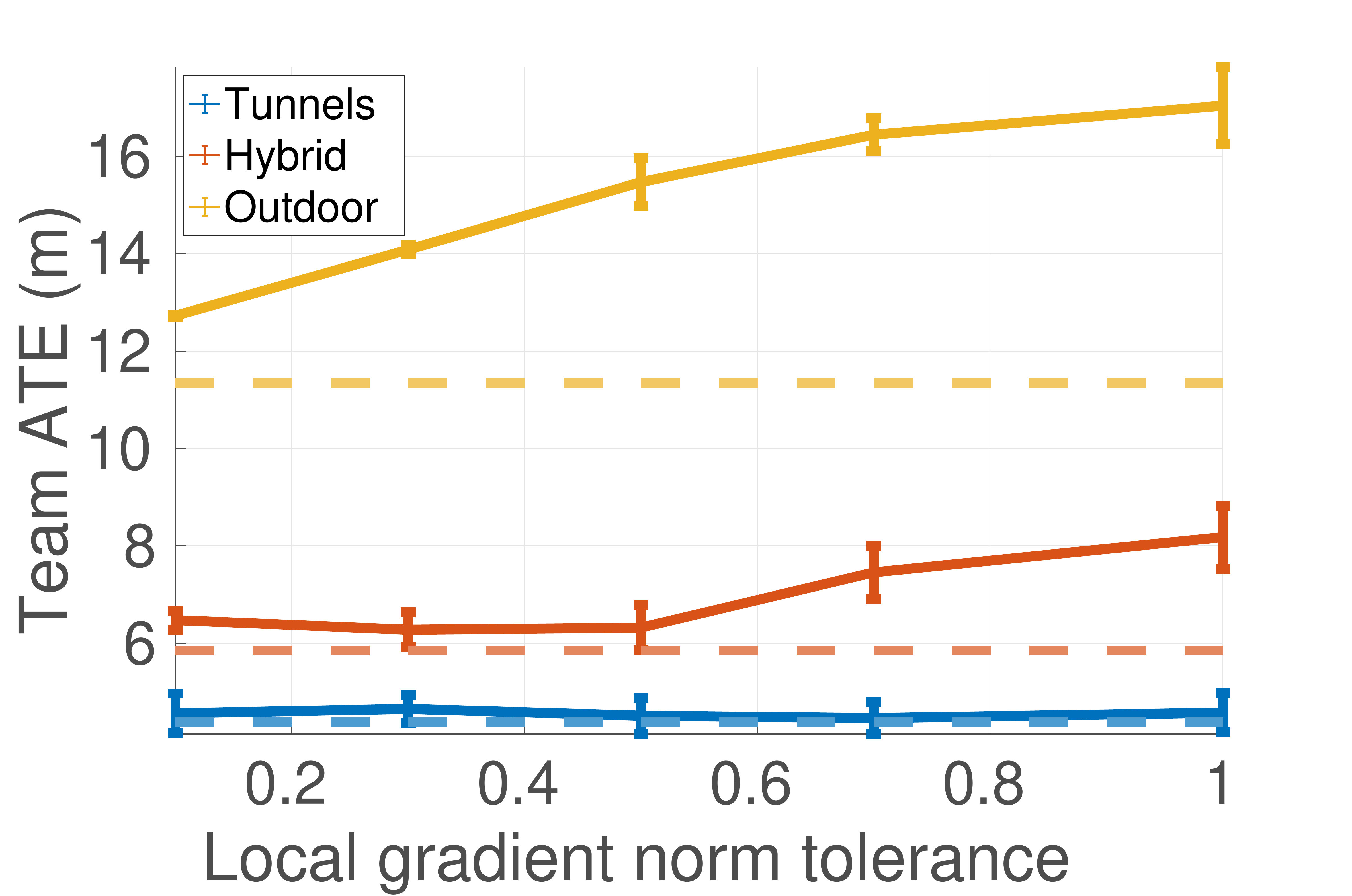}
		\label{fig:ablation_gradnorm}
	} ~
	\subfloat[$\epsilon_{\text{rel}}$]{%
		\includegraphics[trim=20 0 100 0, clip, width=0.14\textwidth]
		{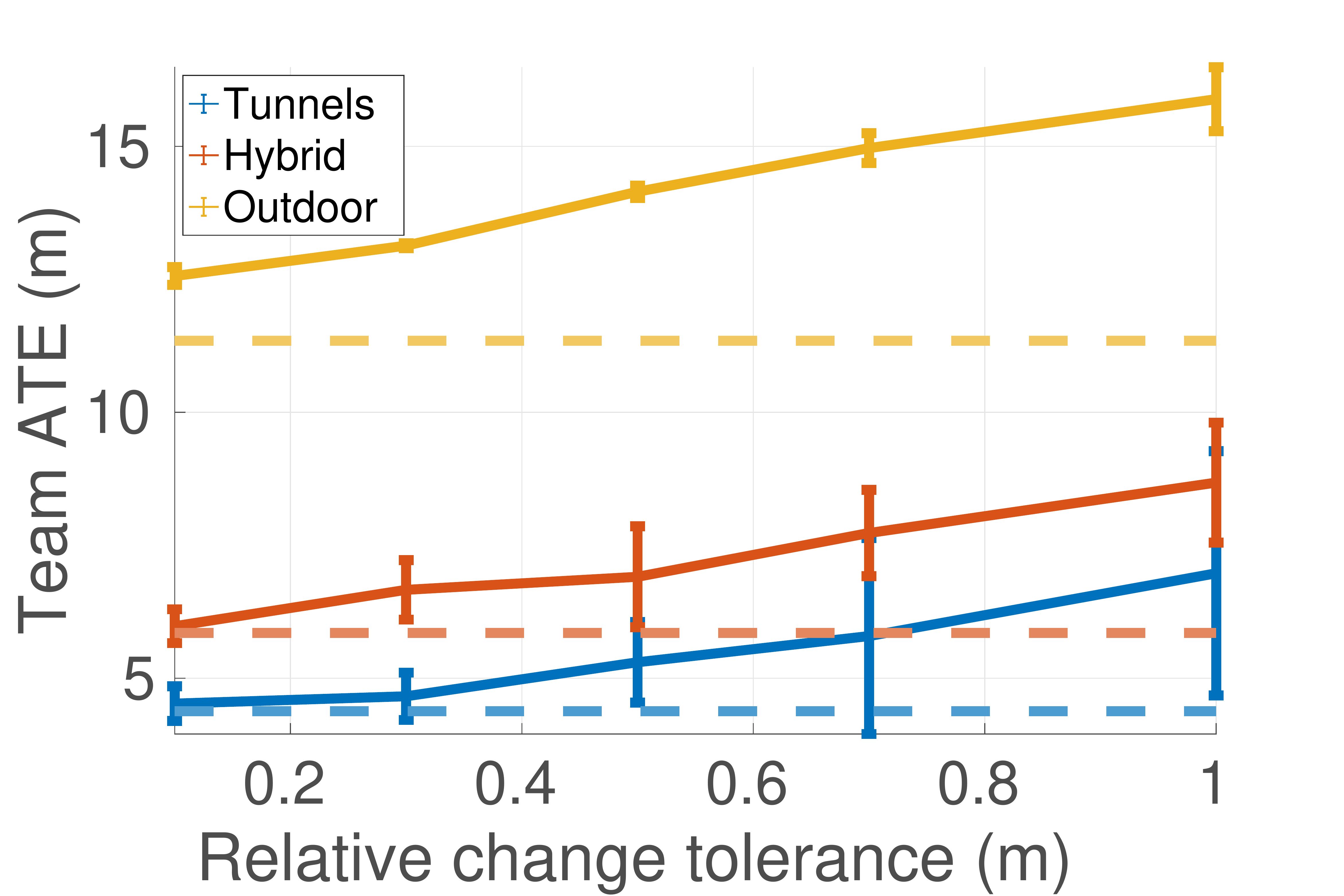}
		\label{fig:ablation_relchange}
	} ~
	\subfloat[$d$]{%
		\includegraphics[trim=20 0 100 0, clip, width=0.14\textwidth]
		{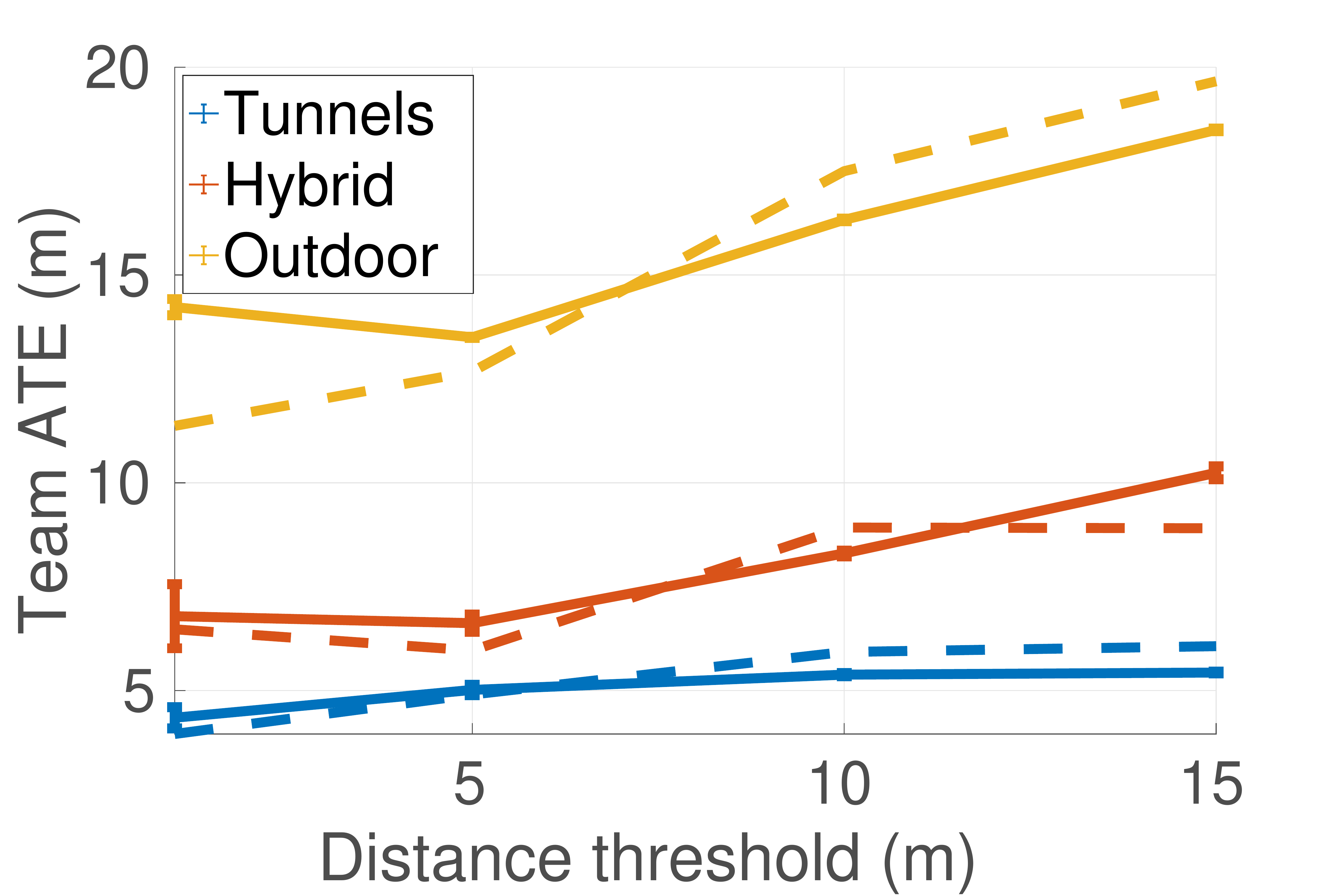}
		\label{fig:ablation_submap}
	} ~
	\caption{
		{	\small 
			Effect on final ATE by varying key parameters:
			(a) local gradient norm threshold $\epsilon_{g}$,
			(b) relative change tolerance threshold $\epsilon_{\text{rel}}$,
			and (c) pose graph coarsening threshold $d$.
		}}
	\label{fig:ablation}
	\vspace{-6mm}
\end{figure}

Fig.~\ref{fig:ablation_gradnorm}-\ref{fig:ablation_relchange} show the impact on ATE as we vary the PGO convergence parameters 
$\epsilon_{g}$ and $\epsilon_{\text{rel}}$. 
For each dataset, we also show the reference ATE (constant dashed line) obtained from a centralized solver that solves GNC to full convergence.
As expected, using smaller values (\ie tighter termination conditions) helps the distributed back-end achieve similar accuracy as the reference solution. 
{However, smaller values also lead to increased number of iterations; 
	for example, decreasing $\epsilon_{\text{rel}}$ from $1.0$ to $0.1$ leads to $51$\%-$180$\% more iterations across the three datasets.}
Larger values lead to worse solutions as optimization terminates before correcting all errors.
Among the three datasets, \Outdoor is the most sensitive to parameter changes due to poor conditioning caused by sparse loop closures.
Meanwhile, \Tunnels is less sensitive to the choice of $\epsilon_{g}$, again due to the abundance of loop closures (\Cref{tab:summary}) that helps robots improve their estimates even with a loose convergence parameter.

Fig.~\ref{fig:ablation_submap} shows the effect of changing the distance threshold $d$ for pose graph coarsening.
The dashed lines again show the ATEs achieved by the centralized solver.
Note that the dashed line is no longer constant because each value of $d$ corresponds to a different PGO problem.
{Generally, decreasing $d$ (\ie increasing the pose graph resolution) yields better ATE.
On the other hand, larger values of $d$ significantly improve the efficiency of the back-end: across the three datasets, increasing $d$ from $1$m to $5$m decreases the number of iterations by $38$\%-$54$\%.}
We observe slightly different trends across datasets, 
where \Outdoor is once again more sensitive to parameter changes.
Lastly, on the \Outdoor dataset we also observe a small increase in ATE towards smaller value of $d$.
This is due to the fact that as $d$ decreases, the pose graph becomes larger and the distributed back-end requires more iterations to achieve better accuracy.

\section{Live Results and Discussions}
\label{sec:discussion}

In this section, we discuss key lessons learned
from live field tests of \kimeraMulti and quantitative results in \Cref{sec:experiments}.
We summarize the capabilities and limitations of our system,
and present challenges and future work towards the resilient deployment of distributed multi-robot SLAM systems.

\begin{figure}[t]
	\centering
	\subfloat[Hybrid (5 out of 8 robots)]{%
		\includegraphics[trim=20 30 20 70, clip, width=0.18\textwidth]
		{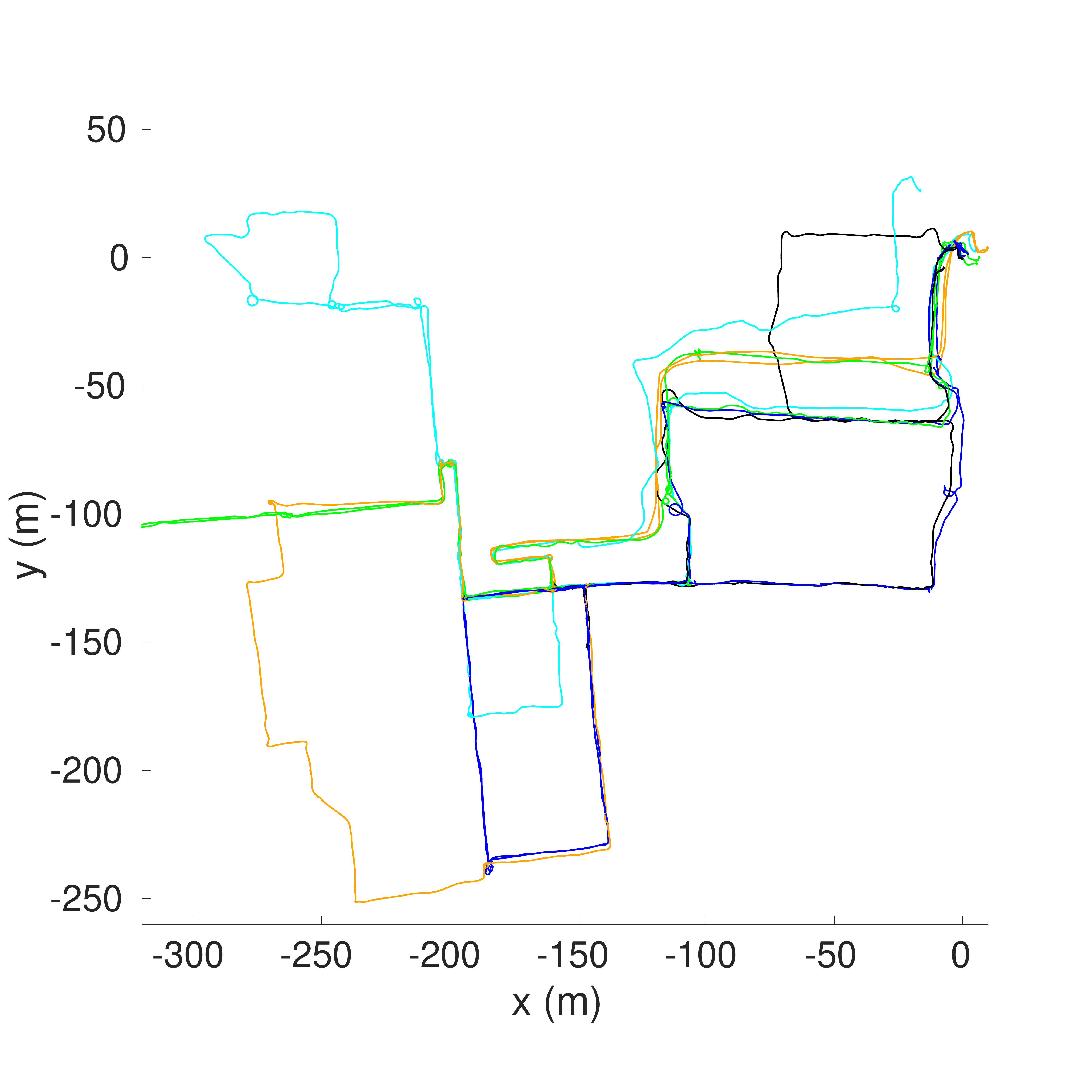}
		\label{fig:1209_test2}
	} ~
	\subfloat[Tunnels (8 robots)]{%
		\includegraphics[trim=20 0 0 70, clip, width=0.18\textwidth]
		{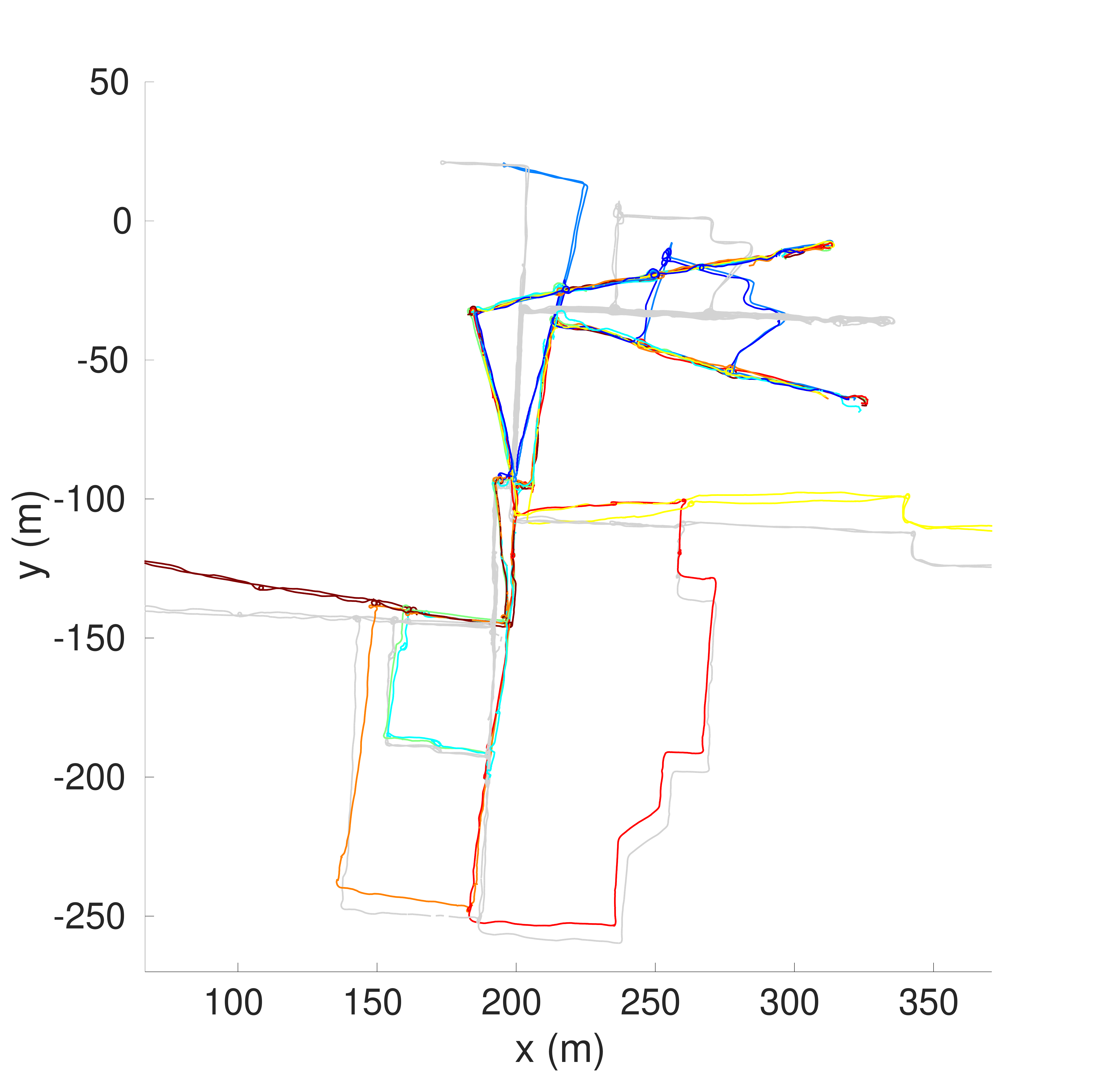}
		\label{fig:1207_test3}
	} 
	\caption{
		{	\small
			Trajectory estimates from example live experiments.
	}}
	\vspace{-5mm}
	\label{fig:live_runs}
\end{figure}

{\bf Resilience to Real-World Failures.}
Failures can happen in unexpected ways during real-world deployments. 
\kimeraMulti is resilient to some of these failure cases, and maintains its core CSLAM capability even if a subset of the robots experience total failures.
Fig.~\ref{fig:1209_test2} shows a representative field test result,
in which 8 robots initially explored a mixed indoor and outdoor scene similar to the \Hybrid dataset,
but 3 robots experienced hardware failures (2 robots ran out of battery and 1 robot's camera went offline) 
during the mission.
Our system was able to adapt to the situation and obtain reasonable trajectory estimates for the remaining 5 robots.
In general, resilience to unexpected failures is crucial for the reliable deployment of CSLAM systems.

{\bf Distributed Front-End (Loop Closure Detection).}
Oftentimes, the accuracy of trajectory estimation depends crucially on detecting sufficient loop closures.
As shown in our experiments, by parallelizing loop closure detection among robots, 
the distributed front-end in \kimeraMulti is able to detect loop closures faster than the centralized baseline.
Nevertheless, our front-end is subject to two limitations.
First, the BoW-based place recognition is sensitive to the type of scenes, and often detects too many matches indoor and not enough outdoor (despite having trained the BoW vocabulary using both indoor and outdoor data).
This makes setting paramaters such as the similarity threshold $\alpha$ a challenging task, 
and using incorrect values could lead to significantly degraded performance. 

Second, \kimeraMulti only performs visual loop closure detection, which is sensitive to changes in the viewpoint.
This negatively impacts the performance on the \Outdoor dataset, where some robots visit the same locations in opposite directions (\eg right part of Fig.~\ref{fig:campus_outdoor}),  
and consequently no loop closure is detected and no trajectory correction occurs.
View dependence is a well-known issue of vision-based loop closure methods.
Further improvement in this module or incorporation of other approaches
(\eg hierarchical and object-based loop closures \cite{Hughes22rss-hydra} and utilizing additional hardware such as ultra-wideband sensors)
would be needed for \kimeraMulti to be resilient to different missions and trajectory plans.

{\bf Distributed Back-End (Pose Graph Optimization).}
Our results demonstrate that our back-end based on distributed GNC \cite{Tian21tro-KimeraMulti} is able to achieve comparable accuracy as a centralized solver, 
and offers additional flexibility by enabling optimization within clusters of connected robots.
Nevertheless, compared to our previous work \cite{Tian21tro-KimeraMulti} that mostly evaluates on small to medium scale problems with 3 robots,
our latest large-scale experiment with 8 robots shows that the accuracy of distributed GNC 
comes at the cost of increased runtime (\Cref{tab:summary}).
{This is because on these larger problems, the distributed PGO solver requires more iterations to solve the intermediate optimization problems within GNC.}
Using a loose termination condition in the distributed solver could cause optimization to stop before correcting all errors, or even cause the rejection of critical inlier loop closures.  
Fig.~\ref{fig:1207_test3} shows an example failure case from a live field test for \Tunnels, where we used a loose relative change threshold of $\epsilon_{\text{rel}} = 1$m.
The result showed a significant error between the estimated trajectories (colored based on robots) and reference solution (colored in gray) on the top right part of the map.
The worse performance is also consistent with the parameter sensitivity analysis presented in Section~\ref{sec:experiments_parameters}.
In general, striking a good balance between optimization time and estimation accuracy can be difficult,
and more work is needed to further improve the speed and scalability of distributed back-ends.

{\bf Communication.}
The communication module in \kimeraMulti is able to handle realistic network conditions with delays and message drops.
However, in scenarios such as the \Tunnels dataset where there are many loop closures,
the front-end sometimes uses most of the available bandwidth for exchanging visual keypoints and descriptors, causing network congestion that interferes with the distributed back-end.
To address this issue, recent methods that prioritize and impose a communication budget on inter-robot loop closure detection \cite{tian19resource,Chang22ral-LAMP2,Lajoie23Swarm} would be helpful.
Another potential solution is to implement \emph{system-level} prioritization that allocates communication resources
to different modules based on their current importance and the network condition.

\section{Conclusion}
\label{sec:conclusion}
This paper describes experimental efforts towards deploying
\kimeraMulti in a large-scale urban environment.
As a result of our field tests, we also contributed new challenging multi-robot CSLAM datasets with accurate reference trajectories and maps.
Our datasets feature many robots traversing diverse indoor and outdoor scenes, facing additional challenges such as severe visual ambiguities and dynamic objects.
We perform extensive quantitative analysis including comparisons of \kimeraMulti against a centralized baseline under simulated communication disruptions.
Our results validate the accuracy and resilience of our system,
identify factors affecting its performance,
and come with a discussion about lessons learned.

{\smaller 
\bibliographystyle{IEEEtran}
\bibliography{include/refs.bib,include/myRefs.bib,include/cslam.bib}
}

\end{document}